\documentclass{article}
\usepackage{etoolbox}
\PassOptionsToPackage{numbers, compress}{natbib}
\usepackage[main, final]{neurips_2026}

\usepackage[utf8]{inputenc}
\usepackage[T1]{fontenc}
\usepackage{hyperref}
\hypersetup{hidelinks}
\usepackage{url}
\usepackage{booktabs}
\usepackage{amsfonts}
\usepackage{amsmath,amssymb}
\usepackage{nicefrac}
\usepackage{microtype}
\usepackage{xcolor}
\usepackage{graphicx}
\usepackage{algorithm}
\usepackage{algpseudocode}
\usepackage{multirow}
\usepackage{placeins}
\usepackage{xspace}
\graphicspath{{figures/}}
\usepackage{booktabs}
\usepackage{multirow}
\usepackage{colortbl}
\usepackage{xcolor}
\usepackage{graphicx}
\usepackage{caption}
\usepackage{wrapfig}
\usepackage{authblk}
\usepackage{xspace}
\usepackage{array}

\usepackage[capitalize]{cleveref}

\newcommand{\method}{GMO-E\textsuperscript{2}DIT\xspace}
\newcommand{\bench}{EComEditBench\xspace}
\newcommand{\op}[1]{\texttt{#1}}
\newcolumntype{L}[1]{>{\raggedright\arraybackslash}m{#1}}

\title{GMO-E\textsuperscript{2}DIT: Grounded Multi-Operation Editing for E-Commerce Images}

\author{%
  \textbf{Zipeng Guo}$^{1}$\thanks{Equal contribution.} ,
  \textbf{Xiaoan Liu}$^{1,2,\ast}$ ,
  \textbf{Lichen Ma}$^{1,3,\ast}$\thanks{Project Lead.} ,
  \textbf{Cheng Wang}$^{4}$ ,
  \textbf{Yu He}$^{1}$ ,
  \textbf{Xiaolong Fu}$^{1}$ ,
  \textbf{Jingling Fu}$^{1}$, 
  \textbf{Xinyuan Shan}$^{1}$ ,
  \textbf{Shaojie Guo}$^{1}$ ,
  \textbf{Luohang Liu}$^{1}$ ,
  \textbf{Junshi Huang}$^{1}$\thanks{Corresponding Author.} ,
  \textbf{Yan Li}$^{1}$ \\
  $^1$JD.com,
  $^2$Wuhan University,
  $^3$Xi'an Jiaotong University,
  $^4$Sun Yat-sen University \\
  \texttt{\{guozp8888, malichen2020, junshi.huang\}@gmail.com}} 

\begin{document}
\maketitle

\begin{abstract}
Real-world e-commerce image editing often requires multiple, localized, and auditable operations rather than global restyling. This compositional nature poses a dual challenge: models must precisely apply all requested edits to the correct regions while preserving unmodified content, even under ambiguous instructions. Existing one-shot editors conflate intent resolution, spatial grounding, and synthesis into a single step, frequently resulting in partial execution failures, which is unacceptable for commercial scenarios. To address this, we introduce \method, an agentic editing framework that couples a Vision-Language Model (VLM) with a mask-conditioned image editor to tackle structured multi-turn task completion. Given an underspecified instruction, the VLM agent constructs a region-grounded edit agenda, effectively decoupling cognitive reasoning from generative rendering. The framework then executes sub-programs via operation-aware masks and references, utilizing a reflection-driven loop to inspect intermediate results and determine the subsequent state. This iterative mechanism reliably preserves safe partial progress, retries unfinished operations, and recovers from errors. Furthermore, we develop a unified data pipeline providing aligned supervision for planning, execution, and reflection, alongside \bench, a comprehensive benchmark for instruction-driven evaluation. Extensive experiments demonstrate that \method achieves competitive performance compared to strong closed-source models, yielding superior instruction accuracy and edit fidelity over existing baselines.
\end{abstract}

\begin{figure}[htbp]
  \centering
  \vspace{-2mm}
  \includegraphics[width=0.97\linewidth]{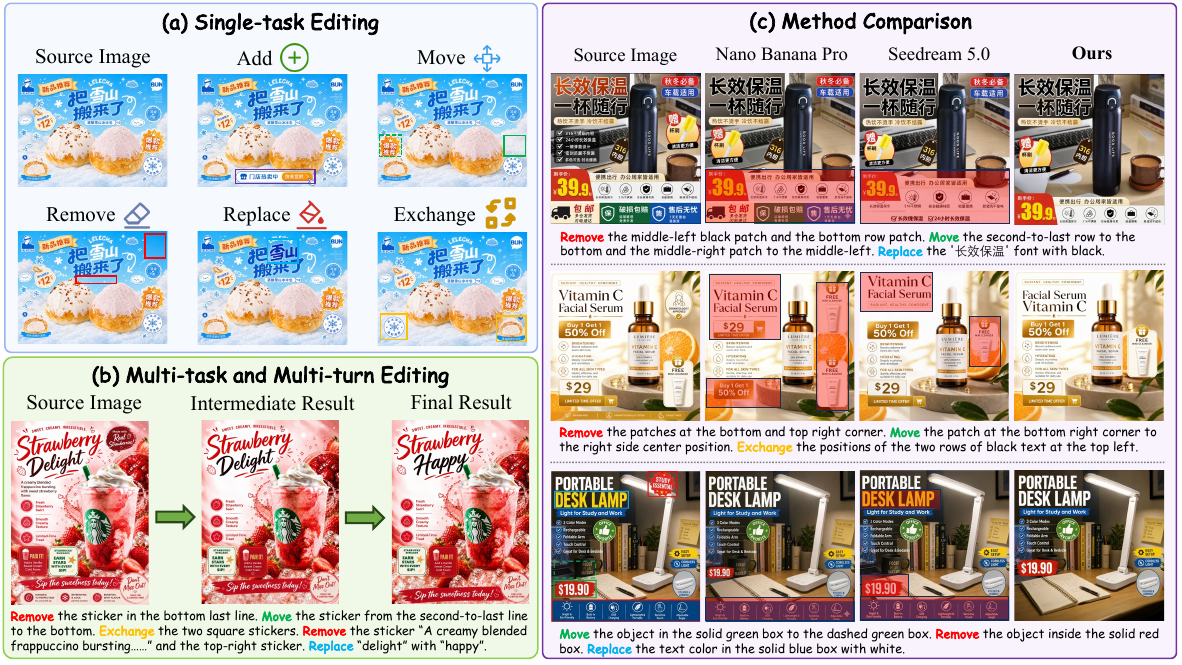}
  \caption{Qualitative demonstration. (a) Single-task editing operations. (b) Complex multi-task and multi-turn editing capabilities. (c) Method comparison demonstrating our approach outperforms strong baselines in executing compositional instructions while preserving unedited regions.}
  \vspace{-6mm}
  \label{fig:teaser_figure}
\end{figure}

\vspace{-2mm}
\section{Introduction}
\vspace{-1mm}
Instruction-driven image editing has become a natural interface for visual content creation, allowing models to perform diverse semantic modifications on input images according to free-form user instructions~\cite{brooks2023instructpix2pix,fu2024mgie,yu2025anyedit,wu2025qwenimage,blackforestlabs2025fluxkontext, xu2025insightedit, zhao2026imageedit, ma2026talk2image,li2025uniworld,zhang2025enabling, zhang2026meta}. For e-commerce product images, however, mere visual plausibility is insufficient. Many visual elements such as logos, price tags, and decorative patches often carry business significance. As shown in Fig.~\ref{fig:teaser_figure} (a), tasks like “move the right-side yellow sticker to the left-side area”, “exchange the bottom-left patch with the bottom-right patch” or “remove the center product sticker and the top-right sticker” require precise, accountable, and auditable local modifications rather than global restyling.

Current one-shot editors struggle to meet these requirements. User instructions are often fuzzy and compositional, referencing ambiguous regions or requesting multiple simultaneous edits. Existing editors typically merge intent resolution, spatial grounding, operation binding, and rendering into a single generation step. Consequently, partial failures are common: some local edits may be correct, while others are missed or erroneously modified. For commercial applications, such partially incorrect outputs are more problematic than incomplete ones. Mask- or box-guided editors\cite{ma2024chargen,yu2025objectmover,wang2025towards,ju2024brushnet,xie2025turbofill,guo2025repainter,ma2026text} provide more control, but practical users rarely supply such annotations, and manual markings can interfere with the product image itself. Although reasoning-guided methods using VLMs and MLLMs can decompose instructions and ground edits~\cite{qu2025replan,fang2025got,bai2026mcie,li2025editthinker,li2026thinkrledit}, they remain largely auxiliary, operating around a fixed generator.

To better tackle the challenges of multi-task, local-operation editing in e-commerce images, we reformulate the problem as multi-operation task completion. The key difficulty lies not only in interpreting natural-language instructions but also in managing a complete editing session until all fine-grained requirements are fulfilled. To this end, we propose \method, a framework in which a VLM acts as both a strategic planner and a reflective judge. Specifically, \method converts underspecified user instructions into a structured edit agenda, explicitly specifying operation types (e.g., \textit{move, add, remove}) and their spatial-semantic constraints, thereby separating cognitive reasoning from generative rendering. This design enables robust multi-turn interaction, where a reflection-driven loop allows the agent to autonomously assess each output and determine the state of each operation, indicating whether it is \textit{success, continue} or \textit{rollback}, and ensuring reliable task decomposition and execution in complex e-commerce scenarios.

To support training and evaluation, we construct a unified e-commerce data pipeline and dataset that provides supervision for planning, execution, and reflection from structured metadata, following the principle that complex visual editing benefits from task-specific data~\cite{bai2026mcie,chen2026posteromni,chen2026macro}. The pipeline extracts editable elements from product and scene images, generates controllable source-target pairs, and provides aligned supervision for the VLM agent and mask-conditioned editor. We further introduce \bench, a benchmark for fuzzy-instruction editing, where the system infers the full edit agenda, and precise-instruction editing, where annotated regions isolate execution quality. The benchmark covers \textit{add, remove, move, exchange, replace} and mixed multi-operation compositions, evaluating instruction accuracy, region fidelity, and content preservation.

In summary, our contributions are threefold. (1) We formalize e-commerce editing as grounded multi-operation task completion, serving as a  paradigm for multi-task, local-operation editing in e-commerce images and providing a practical framework for handling complex, compositional edits. (2) We propose \method, a VLM-based agent that integrates planning and reflection for multi-turn editing, interacting with a mask-conditioned editor to achieve robust and controllable results. (3) We develop a unified data pipeline and \bench for systematic training and evaluation, demonstrating substantial improvements in instruction adherence, region fidelity, content preservation, and reliability across multi-operation editing tasks.

\vspace{-1mm}
\section{Related Work}
\vspace{-1mm}
\label{sec:related}

\textbf{Instruction-based image editing.} Instruction-based image editing aims to modify an input image according to natural-language instructions while preserving irrelevant content. Early diffusion-based methods edit images through attention control, region discovery, or instruction-to-image supervision~\cite{brooks2023instructpix2pix, hertz2022prompt,couairon2023diffedit}. Later datasets and MLLM-assisted editors improve instruction diversity and semantic understanding~\cite{fu2024mgie,yu2025anyedit, zhang2023magicbrush}, while recent unified and in-context models further support open-ended editing, multimodal generation, and reference-conditioned synthesis~\cite{wu2025qwenimage,blackforestlabs2025fluxkontext,zhang2025icedit,liu2025step1xedit,wu2025omnigen2,deng2025bagel,blackforestlabs2025flux2}. Beyond general-purpose editing, specialized tasks such as text-centric editing and poster creation pose unique challenges in maintaining glyph fidelity, layout consistency, and entity preservation~\cite{chen2026posteromni, zhang2026weedit}. E-commerce image editing further requires precise local manipulation of commercial elements, including badges, labels, stickers, and product patches. Accordingly, \method formulates e-commerce editing as grounded multi-operation execution, where executable edit programs explicitly bind operations, regions, and attributes to guide mask-conditioned local rendering.

\textbf{Grounded planning for editing.} Reliable grounding is essential for complex editing. Open-vocabulary detectors and promptable segmentation models provide region-level priors~\cite{liu2024groundingdino,kirillov2023sam}, and recent VLM/MLLM-guided methods use such priors to localize targets, decompose instructions, or inject spatial constraints into diffusion editors~\cite{huang2024smartedit,zhou2025fireedit,yeh2025xplanner}. For example, MCIE introduces spatial- and background-aware attention for complex instruction editing~\cite{bai2026mcie}, RePlan employs a VLM planner to produce region-aligned guidance for diffusion execution~\cite{qu2025replan}, and GoT formulates semantic-spatial reasoning chains with coordinates for generation and editing~\cite{fang2025got}. These methods demonstrate the value of planning before synthesis, yet they mainly focus on identifying where to edit. E-commerce editing additionally requires binding operations to roles, e.g., moving a patch from a source to a target, exchanging two elements, or inserting an external reference. \method addresses this by representing each edit as a typed program entry with source regions, target regions, attributes, and optional patch references, which are compiled into operation-aware masks for execution.

\textbf{Reward feedback editing.} Feedback signals are increasingly used to improve instruction following in image generation and editing. VLM-as-judge protocols provide scalable evaluation, while reward models and RL-based post-training optimize instruction fidelity, visual quality, and content preservation~\cite{zheng2023judging,wu2025editreward,zhao2026firm}. Recent reasoning-centric editors further introduce explicit critique or reflection. EditThinker iteratively critiques editing results and refines prompts for subsequent attempts~\cite{li2025editthinker}, and ThinkRL-Edit expands RL exploration from denoising to reasoning trajectories through planning, reflection, and checklist-style rewards~\cite{li2026thinkrledit}. Such methods highlight the importance of deliberation, but their feedback is usually prompt-level or globally scored. In contrast, the reflection module in \method is program-level: it classifies each intermediate result as \op{success}, \op{continue}, or \op{rollback}. This allows the system to keep correct partial edits, retry only missing operations, and discard outputs that damage the product image, enabling reliable multi-operation execution.

\begin{figure}[htbp]
  \centering
  \includegraphics[width=0.97\linewidth]{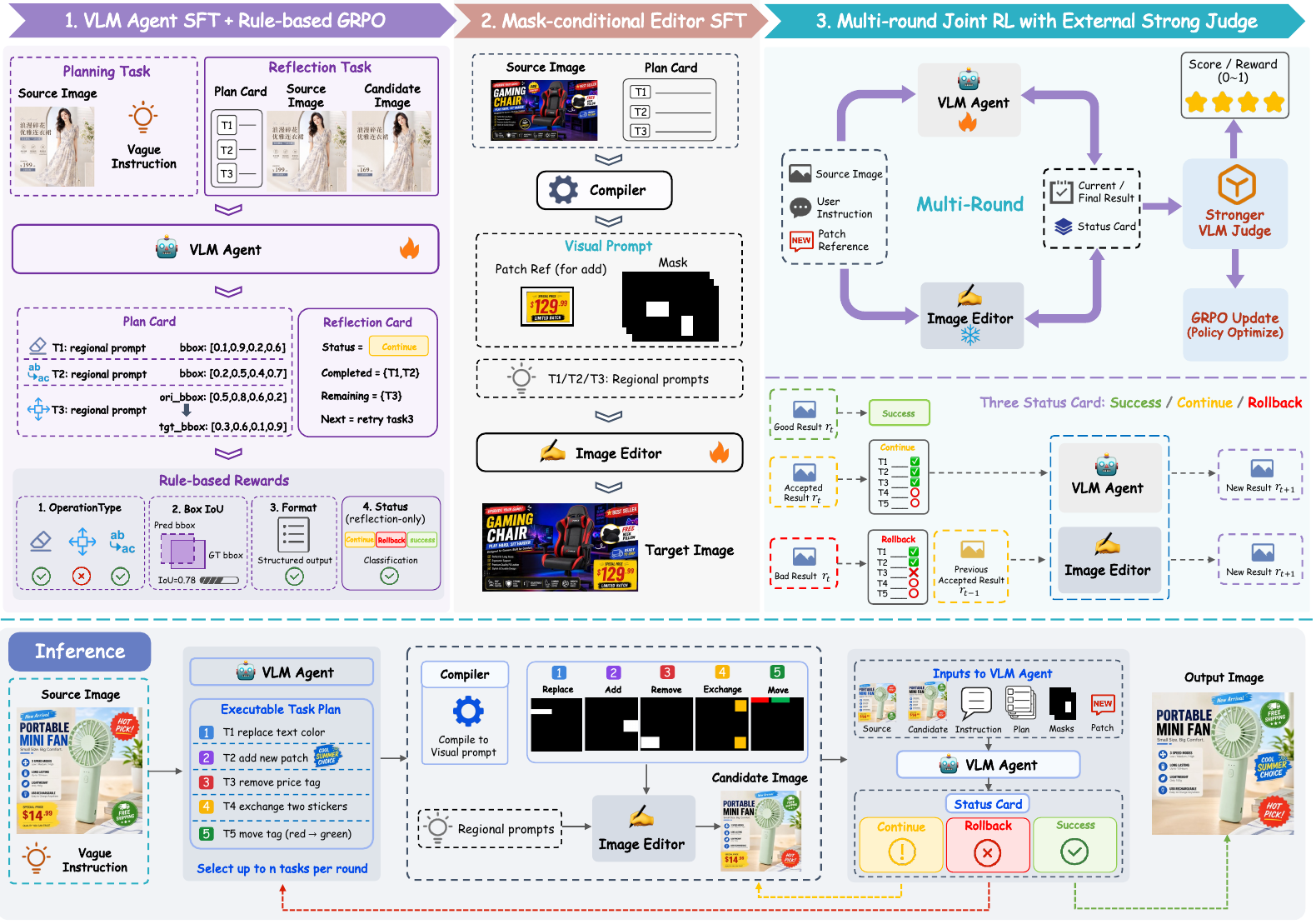} 
  \caption{Overview of our framework. Training involves three stages: 1) SFT and rule-based GRPO for the VLM agent; 2) SFT for the mask-conditioned editor; and 3) multi-round joint RL guided by an external VLM judge. During inference, the agent and editor interact in a closed loop, iteratively executing masked visual prompts and refining results until the instruction is complete.}
  \vspace{-5mm}
  \label{fig:method_overview}
\end{figure}

\section{Method}
\label{sec:method}

\subsection{Problem definition}
Given a source e-commerce image $x\in[0,1]^{H\times W\times 3}$ and a natural-language editing instruction $u$, our goal is to produce an edited image $y$ that completes all requested local operations while preserving non-target regions. We represent the latent intent as a structured \emph{edit agenda},
\begin{equation}
\mathcal{U}=\{p_i\}_{i=1}^{K},\qquad
p_i=(\tau_i,b_i^{\mathrm{src}},b_i^{\mathrm{dst}},a_i,h_i),
\end{equation}
where $i$ denotes the task order, $\tau_i$ the operation type, $b_i^{\mathrm{src}}$ and $b_i^{\mathrm{dst}}$ optional source and target boxes, $a_i$ operation-specific attributes, and $h_i$ a short textual hint. The operations include addition, removal, movement, exchange, and local replacements. This role-aware design separates reasoning from rendering: the VLM decides \emph{what} and \emph{where} to edit, while the editor handles localized synthesis.

\subsection{Agent--Editor Framework}
\label{sec:agent_editor_framework}

Our framework contains a VLM editing agent $\mathcal{G}_{\phi}$ and a mask-conditioned image editor $\mathcal{E}_{\theta}$. The VLM agent works in two modes: \emph{planning} and \emph{reflection}. In the planning mode, it converts the fuzzy instruction $u$, together with the source image $x\in[0,1]^{H\times W\times 3}$, into a structured edit agenda,
\begin{equation}
\hat{\mathcal{U}}_0=\mathcal{G}_{\phi}^{\mathrm{plan}}(x_0,u).
\end{equation}
At editing round $t$, a scheduler selects a small sub-program $\mathcal{Q}_t\subseteq \hat{\mathcal{U}}_{t-1}$ from the remaining agenda. A deterministic compiler $\mathcal{C}$ converts $\mathcal{Q}_t$ into a canonical prompt $\tilde{u}_t$, a small set of operation-aware masks $\mathcal{M}_t$, and optional patch references $\mathcal{A}_t$, based on which the intermediate image $\hat{y}_t$ is generated by editor $\mathcal{E}_{\theta}$, 
\begin{equation}
(\tilde{u}_t,\mathcal{M}_t,\mathcal{A}_t)=\mathcal{C}(x_t,u,\mathcal{Q}_t),
\qquad
\hat{y}_t=\mathcal{E}_{\theta}(x_t,\tilde{u}_t,\mathcal{M}_t,\mathcal{A}_t).
\end{equation}

After the editor generates an intermediate result, the same VLM agent switches to reflection mode:
\begin{equation}
(\hat{f}_t, \hat{s}_t)=\mathcal{G}_{\phi}^{\mathrm{reflect}}(x_t,\hat{y}_t,\mathcal{U}_{t-1},\mathcal{Q}_t),
\qquad \hat{s}_t\in\{\op{success},\op{continue},\op{rollback}\}.
\end{equation}
\vspace{-2mm}
\begin{equation}
\label{eq:state_transition}
(x_{t+1},\hat{\mathcal{U}}_t)=
\begin{cases}
(\hat{y}_t,\varnothing), & \hat{s}_t=\op{success},\\
(\hat{y}_t,\textsc{Remaining}(\hat{f}_t)), & \hat{s}_t=\op{continue},\\
(x_t,\textsc{Replan}(\hat{f}_t,\hat{\mathcal{U}}_{t-1},u)), & \hat{s}_t=\op{rollback}.
\end{cases}
\end{equation}
The reflection output evaluates the result of each editing round and assigns one of three states: \op{success}, \op{continue}, or \op{rollback}. If all tasks are successfully completed (\op{success}), the system stops, as no further edits are required. If some tasks remain but no critical errors occurred (\op{continue}), the agent continues with the remaining tasks. If significant errors occur (\op{rollback}), the system rolls back to the input or last safe image and retries with a revised sub-program. This mechanism ensures multi-turn editing preserves safe progress while preventing errors from propagating.


\subsection{Training and Inference}
\label{sec:training_inference}

As shown in Fig.~\ref{fig:method_overview}, our framework trains a VLM agent and a mask-conditioned editor for reliable multi-turn, multi-operation editing. The agent is first trained to generate structured edit agendas and diagnose intermediate states, while the editor learns to execute these programs with explicit visual grounding. In a joint training stage, the agent interacts with the frozen editor and receives rewards for both plan accuracy and reflection, learning when to continue or roll back. At test time, the system iteratively applies the agenda with reflection-guided retries to achieve robust, multi-turn editing.

\textbf{Stage 1: Supervised and Reinforcement Training of the Agent.}
We first train the VLM agent $\mathcal{G}_{\phi}$ with supervised fine-tuning on both planning and reflection data. Planning samples map an image-instruction pair $(x,u)$ to a edit agenda $\mathcal{U}$, while reflection samples map an intermediate editing state $(x_t,\hat{y}_t,\mathcal{U}_{t-1},\mathcal{Q}_t)$ to a diagnosis $f_t$ and a state label $s_t$. The SFT objective is
\begin{equation}
\mathcal{L}_{\mathrm{agent\text{-}sft}}(\phi)
=-\mathbb{E}_{(q,o)\sim\mathcal{D}_{\mathrm{agent}}}
\sum_{l=1}^{|o|}\log \pi_{\phi}(o_l\mid q,o_{<l}).
\end{equation}

To further improve structured planning and state diagnosis, we apply GRPO to the VLM agent. For each query $q$, the current policy samples a group of responses $\{o_g\}_{g=1}^{G}$, and the advantage is normalized within the group. The optimization objective follows
\begin{equation}
\mathcal{J}_{\mathrm{GRPO}}(\phi)
=\mathbb{E}_{q,\{o_g\}}
\frac{1}{G}\sum_{g=1}^{G}
\min\left(
r_gA_g,
\mathrm{clip}(r_g,1-\epsilon,1+\epsilon)A_g
\right)
-\beta D_{\mathrm{KL}}(\pi_{\phi}\|\pi_{\mathrm{ref}}),
\end{equation}
where $r_g=\pi_{\phi}(o_g\mid q)/\pi_{\mathrm{old}}(o_g\mid q)$ and $\pi_{\mathrm{ref}}$ is the reference policy. Notably, we define two groups of verifiable rewards. For planning, the reward evaluates whether the predicted agenda is properly structured, whether the actions follow the annotated order and operation types, and whether the predicted boxes align with the target boxes. For reflection, the reward assesses whether the predicted result $f_t$ is correctly formatted, whether the state label $\hat{s}_t$ corresponds to \op{success}, \op{continue}, or \op{rollback}, and whether the subsequent tasks after the current state match the ground truth.

\textbf{Stage 2: Training the Mask-Conditioned Editor.}
The editor $\mathcal{E}_{\theta}$ is trained to execute localized programs produced by the agent. Each training sample contains the current image $x_t$, the compiled instruction $\tilde{u}_t$, operation-aware masks $\mathcal{M}_t$, optional patch references $\mathcal{A}_t$, and the target edited image $y_t$. By providing visual grounding and role information, the editor is relieved from inferring fuzzy hint and can focus entirely on local synthesis. For a flow-matching editor, the training objective is
\begin{equation}
\mathcal{L}_{\mathrm{edit}}
=
\mathbb{E}\left[
\left\|
v_{\theta}(z_t,t,x_t,\tilde{u}_t,\mathcal{M}_t,\mathcal{A}_t)-v^{\star}
\right\|_2^2
\right],
\end{equation}
where $z_t$ is the noisy latent state and $v^{\star}$ is the target velocity. Diffusion-based editors can use the analogous noise-prediction objective. Since region roles are expressed by masks and patch references, the editor can focus on faithful local rendering, content preservation, and patch consistency.

\textbf{Stage 3: Joint Reinforcement Training with Editor Feedback.}
Finally, the agent is trained in the multi-round editing loop while keeping the editor $\mathcal{E}_{\theta}$ frozen. For each instruction, the policy samples a group of trajectories. In each trajectory, $\mathcal{G}_{\phi}$ first predicts $\hat{\mathcal{U}}_0$; the scheduler selects $\mathcal{Q}_t$; the frozen editor generates $\hat{y}_t$; and the updated context $(x_t,\hat{y}_t,\mathcal{U}_{t-1},\mathcal{Q}_t)$ is fed back to the agent for reflection. If $\hat{s}_t=\op{continue}$, the next round starts from $\hat{y}_t$ and the remaining tasks in $\hat{f}_t$; if $\hat{s}_t=\op{rollback}$, the corrupted result is discarded and the next round restarts from the previous safe image.

The closed-loop reward integrates verifiable reward with model-based evaluation. During planning turns, the agent is rewarded for maintaining the correct structure, action order and types, and bounding-box alignment as established in Stage 1. During reflection turns, a stronger VLM judge evaluates both the edited image and the subsequent tasks selected by the agent. Notably, premature \op{success} decisions on incomplete or incorrect images are heavily penalized to prevent early termination. Rewards are assigned to the token spans responsible for generating the corresponding planning or reflection outputs, and the sampled trajectories are optimized using the GRPO objective. This stage trains the agent not only to generate accurate plans but also to determine when it is safe to continue from an intermediate result and when to roll back, thereby enabling reliable multi-turn execution.

\textbf{Inference.} At test time, the same loop is run for at most $T$ rounds. Starting from $x_0=x$, the system repeatedly updates $(x_{t+1},\hat{\mathcal{U}}_t)$ according to Eq.~\eqref{eq:state_transition}. Execution terminates when $\hat{s}_t=\op{success}$. Otherwise, it continues from $\hat{y}_t$ for recoverable unfinished operations or rolls back to $x_t$ for harmful failures. In this way, the structured agenda provides multi-operation capability, and reflection-guided retry provides multi-turn reliability.

\section{Dataset Construction and Benchmark}

To enhance complex instruction following, we propose a data synthesis pipeline focused on compositional operations, spatial grounding, and iterative feedback. Moving beyond isolated edits, our approach simulates multifaceted e-commerce workflows by integrating tasks such as patch insertion, region relocation, and attribute modification (e.g., text, color, and style). This provides rigorous supervision for the model's spatial localization, generative fidelity, and instruction adherence.

\begin{figure}[htbp]
  \centering
  \includegraphics[width=0.97\linewidth]{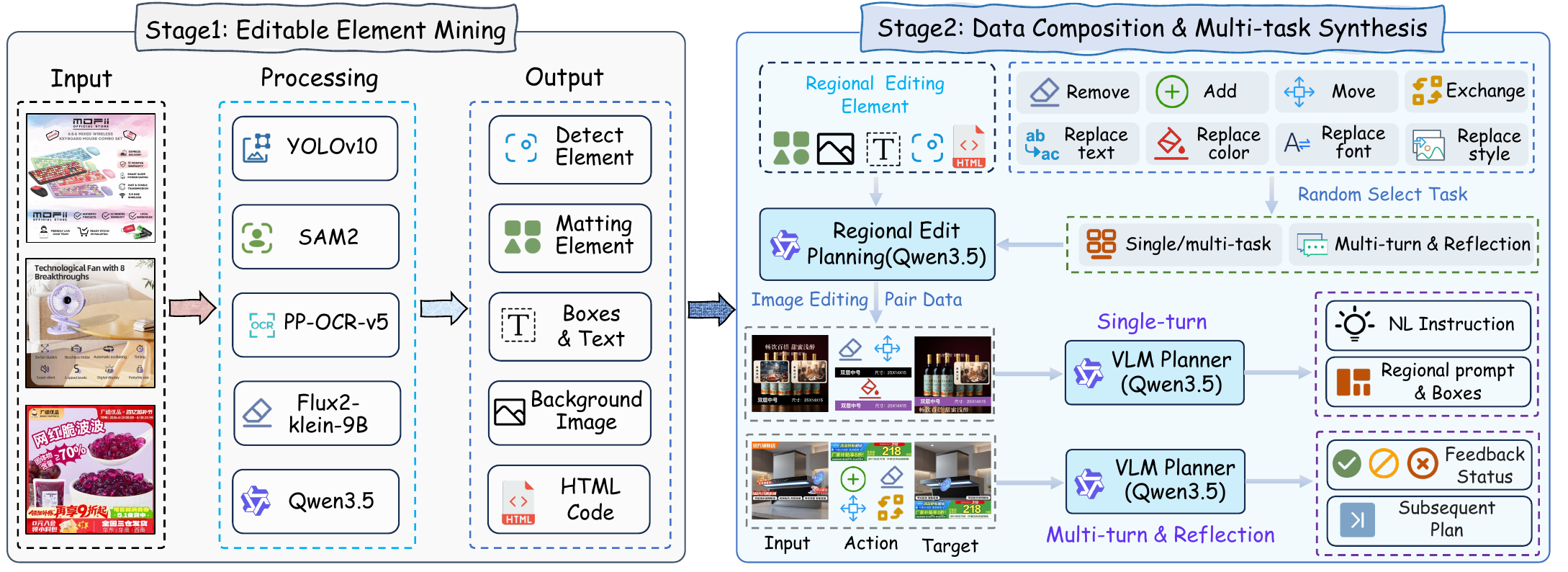} 
  \vspace{-2pt}
  \caption{Overview of the data synthesis pipeline. Stage 1 extracts structured visual and textual assets, while Stage 2 recomposes them to synthesize grounded multi-task and multi-turn training pairs.}
  \vspace{-5mm}
  \label{fig:data_pipeline_overview}
\end{figure}


\subsection{Editable Element Mining}
As illustrated in Fig.~\ref{fig:data_pipeline_overview}, the pipeline accepts a diverse array of e-commerce posters as input. These images are characterized by intricate compositional layouts, integrating multi-layered visual and textual components.
We utilize a sophisticated multi-stage processing pipeline to decompose each input image into its constituent semantic layers and structure.
The pipeline generates a comprehensive suite of aligned, editable representations for each input image. This includes pixel-level masks and spatial bounding boxes, high-fidelity matted assets (e.g., products and icons), structured textual metadata with precise coordinates, a semantically-consistent inpainted background, and a hierarchical HTML representation that captures the poster's underlying layout and semantic logic.






\begin{wrapfigure}{r}{0.5\textwidth}
\vspace{-5mm}
  \begin{center}
    \includegraphics[width=0.5\textwidth]{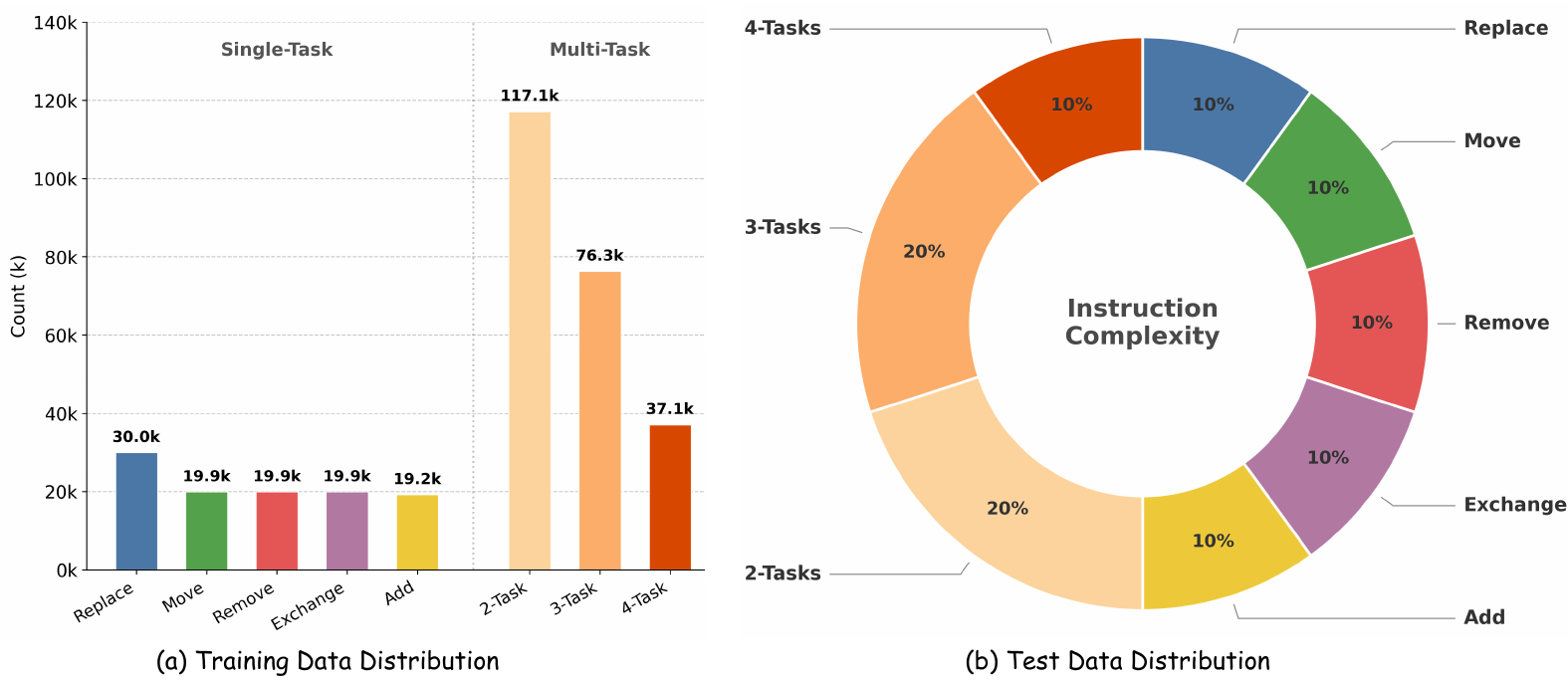}
  \end{center}
  \vspace{-6pt} 
  \caption{Overview of training and test data. (a) Sample counts for single-task and multi-task instructions in training. (b) Evaluation set composition by instruction complexity, highlighting benchmark diversity.}
  \label{fig:data_distributions}
  \vspace{-10pt} 
\end{wrapfigure}

\subsection{Data Composition And Multi-task Synthesis}
\textbf{Unified Synthesis of Image Editing Data.} We synthesize training samples by integrating three core components: the raw product image, a generatively-inpainted background, and structured elements extracted via our perception pipeline. Specifically, we localize semantic components—such as stickers and text blocks—using object detection, followed by alpha-matting to isolate them as high-fidelity, transparent assets. Each asset is stored as a structured tuple encompassing its semantic category, bounding box, and pixel-level mask.

To bridge these discrete assets with complex editing intents, we introduce a dual-mode manipulation strategy. While geometric operations (e.g., addition or relocation) are applied directly, local replacement requires modifying a region's semantic content or style while preserving its spatial and layout integrity. To ensure controllability and interpretability, we adopt HTML as a structured intermediate representation for replacement, effectively decoupling style attributes from spatial constraints. Finally, these elements are programmatically recomposed based on edit programs orchestrated by the Regional Edit Planner (Qwen3.5), yielding precisely aligned source-target pairs for grounded e-commerce editing.

\textbf{Synthesis of Structured Data for Multi-task Planning.} Leveraging the visual differences between source and target pairs, we utilize inverse reasoning to generate training triplets for the VLM planner. Each triplet consists of a natural language instruction, a grounded region mask, and a spatial prompt. We further expand the scope of our planning data by incorporating multi-task regional editing operations, such as \textit{addition}, \textit{removal}, \textit{movement}, \textit{exchange}, and \textit{replacement}. By mapping high-level linguistic directives to these fine-grained spatial constraints, our pipeline establishes a unified data contract that facilitates robust, multi-operation image editing.

\textbf{Synthesis of Iterative Data for Multi-turn Reflection.} Solely training for image generation is insufficient for stable multi-turn editing, as models may partially execute tasks or produce spatial errors. Thus, we construct a reflection dataset to train a feedback model for state diagnosis and decision-making. We categorize feedback into three states: (1) \textit{Success}: All tasks are successfully completed; (2) \textit{Continue}: Tasks are partially finished without fatal errors, allowing the system to continue with the remaining objectives; and (3) \textit{Rollback}: Significant errors (e.g., misplacement or scaling issues) occur, triggering a rollback to the previous input and task decomposition for a retry.


\subsection{Benchmark}

\textbf{Task Setting.} 
We introduce \bench, a held-out benchmark for grounded multi-operation e-commerce editing. Following recent task-specific benchmark design in editing and long-context generation~\cite{chen2026macro,qu2025replan,zhang2026weedit}, it evaluates not only whether an output looks plausible, but whether all requested local edits are executed on the correct regions while the rest of the product remains intact. Fig.~\ref{fig:data_distributions} showcases the benchmark's composition, which consists of 1,000 samples distributed across five atomic operations: \op{add}, \op{remove}, \op{move}, \op{exchange}, and \op{replace}. The benchmark is balanced between 500 single-operation and 500 multi-operation cases, enabling evaluation of both primitive local edits and compositional editing under multiple region constraints. Additional details are provided in Appendix~\ref{app:benchmark_details}.

\textbf{Instruction protocols.}
We evaluate models under fuzzy and precise instruction protocols. The fuzzy protocol provides only the source image and a natural-language request, testing end-to-end instruction understanding, operation binding, and spatial grounding. The precise protocol provides structured region-level edit descriptions, optionally with insertion patches, to isolate localized editing once grounding is given.

\textbf{VLM-based scoring.}
Following recent benchmarks, we use a fixed VLM-as-a-Judge protocol that evaluates the source image, edited result, and optional references, and assigns three integer scores in $[1,10]$. \textbf{Instruction Accuracy (IA)} measures whether all requested edits are applied to the correct objects and locations. \textbf{Edit Fidelity (EF)} evaluates the visual quality and semantic correctness of edited regions. \textbf{Background Preservation (BP)} measures whether non-target regions remain unchanged. We report both fine-grained task-group scores and overall averages.

\section{Experiment}
\label{sec:experiments}

\textbf{Experimental Setup.} We train \method on our constructed e-commerce editing dataset with supervision for region planning, mask-conditioned editing, and visual reflection. We evaluate on \bench using fuzzy and precise instruction protocols, which respectively test end-to-end instruction understanding and localized editing execution. Following VLM-as-judge practice~\cite{zheng2023judging}, we use Gemini-3 Pro to evaluate each output using the source image, instruction, model result, and optional region or patch references. More details can be found in the appendix~\ref{implementation_details}.

\subsection{Experimental Results}

\textbf{Quantitative Results.}
Table~\ref{tab:main_results_detailed} reports the main comparison under fuzzy- and precise-instruction settings. In the fuzzy setting, \method achieves the best overall IA and EF scores ($8.66$ and $8.25$) and the second-best BP score ($8.87$), substantially outperforming open-source baselines such as Qwen-Image-Edit-2511 ($4.52/4.12/4.91$) and FLUX2-klein-base-9B ($5.16/4.53/6.65$) in IA/EF/BP. The gains are especially clear on role-sensitive operations, including \op{move} ($8.87/8.14/8.56$) and \op{exchange} ($9.07/8.54/9.30$), suggesting that explicit agenda planning and operation-aware masks reduce source--target confusion. \method also leads IA and EF on compositional two- and four-operation cases. Although closed-source models such as Nano Banana Pro better preserve backgrounds, they still lag behind \method in instruction completion and edit fidelity.

\begin{table*}[t]
\centering
\caption{Quantitative results on \bench under fuzzy-instruction and precise-instruction settings. Scores are on a 1--10 scale; higher is better. \textbf{Bold} denotes the best and \underline{underlined} denotes the second-best result.
}
\vspace{-2mm}
\label{tab:main_results_detailed}
\scriptsize
\setlength{\tabcolsep}{1.5pt}
\renewcommand{\arraystretch}{1.2}
\resizebox{\textwidth}{!}{%
\begin{tabular}{c|L{2.8cm}|ccc|ccc|ccc|ccc|ccc|ccc|ccc|ccc|ccc}
\toprule
\multicolumn{1}{c}{} &
\multicolumn{1}{L{2.8cm}}{\multirow[c]{3}{*}{\textbf{Method}}}
& \multicolumn{15}{c|}{\textbf{Single Task}}
& \multicolumn{9}{c|}{\textbf{Multi Task}}
& \multicolumn{3}{c}{\multirow[c]{2}{*}{\textbf{Overall}}} \\
\cmidrule(lr){3-17} \cmidrule(lr){18-26}

\multicolumn{1}{c}{} &
\multicolumn{1}{L{2.8cm}}{}
& \multicolumn{3}{c|}{\textbf{Add}}
& \multicolumn{3}{c|}{\textbf{Replace}}
& \multicolumn{3}{c|}{\textbf{Remove}}
& \multicolumn{3}{c|}{\textbf{Move}}
& \multicolumn{3}{c|}{\textbf{Exchange}}
& \multicolumn{3}{c|}{\textbf{2 Tasks}}
& \multicolumn{3}{c|}{\textbf{3 Tasks}}
& \multicolumn{3}{c|}{\textbf{4 Tasks}}
& \multicolumn{3}{c}{} \\
\cmidrule(lr){3-5} \cmidrule(lr){6-8} \cmidrule(lr){9-11}
\cmidrule(lr){12-14} \cmidrule(lr){15-17}
\cmidrule(lr){18-20} \cmidrule(lr){21-23} \cmidrule(lr){24-26} \cmidrule(lr){27-29}

\multicolumn{1}{c}{} &
\multicolumn{1}{L{2.8cm}}{}
& IA & EF & BP
& IA & EF & BP
& IA & EF & BP
& IA & EF & BP
& IA & EF & BP
& IA & EF & BP
& IA & EF & BP
& IA & EF & BP
& IA & EF & BP \\
\midrule

\multirow{10}{*}{\rotatebox[origin=c]{90}{\textbf{Fuzzy}}}
& \multicolumn{28}{l}{\cellcolor{gray!12}\textbf{Closed-source methods}} \\
& Nano Banana 2
& \textbf{9.69} & \underline{9.55} & 9.75 & 9.04 & 9.01 & 9.57 & \textbf{9.67} & \textbf{9.52} & 8.48 & 8.21 & 7.80 & \underline{8.87} & 7.61 & 7.15 & 8.59 & 8.30 & \underline{7.96} & 8.59 & \textbf{8.27} & \textbf{7.99} & \textbf{8.72} & 7.49 & 7.20 & \underline{8.22} & 8.48 & \underline{8.21} & 8.81 \\

& Nano Banana Pro
& \textbf{9.69} & \textbf{9.73} & \underline{9.93} & \underline{9.19} & \underline{9.14} & \underline{9.80} & 9.32 & 9.16 & \textbf{9.08} & 7.93 & 7.61 & \textbf{9.18} & 7.30 & 6.89 & \underline{8.81} & 7.95 & 7.71 & \textbf{9.05} & 7.55 & 7.33 & \underline{8.37} & 7.19 & 7.18 & \textbf{8.82} & 8.16 & 7.98 & \textbf{9.04} \\

& Seedream 5.0
& 9.22 & 8.80 & 8.46 & \textbf{9.20} & \underline{9.14} & 9.49 & \underline{9.52} & 9.24 & 8.39 & \underline{8.82} & \textbf{8.45} & 8.39 & \underline{8.07} & \underline{7.51} & 8.74 & \underline{8.34} & 7.88 & 8.34 & 7.88 & 7.45 & 7.67 & \underline{7.68} & \underline{7.30} & 7.47 & \underline{8.49} & 8.11 & 8.30 \\

& Wan2.7-Image-Pro
& \underline{9.64} & 9.50 & 9.77 & 9.18 & \textbf{9.15} & 9.64 & 9.48 & \underline{9.29} & 8.71 & 7.22 & 6.88 & 8.52 & 3.74 & 3.61 & 8.25 & 6.88 & 6.65 & 8.55 & 6.21 & 6.04 & 8.34 & 4.98 & 4.92 & 7.94 & 7.04 & 6.87 & 8.66 \\

& \multicolumn{28}{l}{\cellcolor{gray!12}\textbf{Open-source methods}} \\
& FireRed-Image-Edit-1.1
& 1.35 & 1.11 & 1.64 & 6.82 & 5.89 & 4.67 & 8.08 & 7.83 & 4.08 & 4.65 & 3.73 & 3.06 & 2.06 & 1.84 & 2.20 & 2.34 & 1.91 & 1.63 & 1.91 & 1.65 & 1.46 & 1.25 & 1.14 & 1.18 & 3.27 & 2.87 & 2.30 \\

& Qwen-Image-Edit-2511
& 4.88 & 3.85 & 5.29 & 8.09 & 8.14 & 7.63 & 8.42 & 8.07 & 6.03 & 4.90 & 4.49 & 5.17 & 2.41 & 2.21 & 4.48 & 3.87 & 3.43 & 4.04 & 3.21 & 2.75 & 4.33 & 2.31 & 2.12 & 3.72 & 4.52 & 4.12 & 4.91 \\

& FLUX2-klein-base-9B
& 7.61 & 5.40 & 8.17 & 7.34 & 7.10 & 8.38 & 8.97 & 8.79 & 8.20 & 5.49 & 4.49 & 7.80 & 2.02 & 1.91 & 6.78 & 4.68 & 3.97 & 6.01 & 4.16 & 3.70 & 5.58 & 2.51 & 2.24 & 3.96 & 5.16 & 4.53 & 6.65 \\

\rowcolor{green!8}
& \textbf{\method{} (ours)}
& 9.62 & 9.31 & \textbf{9.95} & 9.04 & 9.00 & \textbf{9.88} & 8.96 & 8.83 & \underline{9.06} & \textbf{8.87} & \underline{8.14} & 8.56 & \textbf{9.07} & \textbf{8.54} & \textbf{9.30} & \textbf{8.67} & \textbf{8.13} & \underline{8.74} & \underline{8.00} & \underline{7.47} & 8.12 & \textbf{7.75} & \textbf{7.44} & 8.06 & \textbf{8.66} & \textbf{8.25} & \underline{8.87} \\

\midrule
\multirow{10}{*}{\rotatebox[origin=c]{90}{\textbf{Precise}}}
& \multicolumn{28}{l}{\cellcolor{gray!12}\textbf{Closed-source methods}} \\
& Nano Banana 2
& 9.82 & 9.72 & \textbf{9.92} & 9.09 & \underline{9.01} & \textbf{9.92} & \underline{9.54} & 9.45 & 9.19 & 8.50 & 8.32 & 9.17 & \underline{9.32} & \underline{9.06} & \underline{9.59} & 8.86 & 8.63 & 9.27 & 9.30 & 9.10 & 9.44 & 9.07 & 8.95 & 9.19 & 9.17 & 9.00 & 9.44 \\

& Nano Banana Pro
& \underline{9.84} & \underline{9.79} & 9.83 & \underline{9.16} & 8.98 & 9.46 & 9.47 & 9.43 & 9.45 & 8.17 & 7.93 & \underline{9.27} & 9.26 & 8.98 & 9.47 & 9.10 & 8.91 & 9.55 & \underline{9.50} & \underline{9.28} & \underline{9.72} & 8.71 & 8.58 & 8.87 & 9.18 & 9.01 & \underline{9.49} \\

& Seedream 5.0
& 9.74 & 9.46 & 9.68 & 8.65 & 8.46 & 9.44 & \textbf{9.82} & \textbf{9.71} & \underline{9.57} & \underline{8.79} & \underline{8.50} & 9.15 & 9.02 & 8.80 & 9.53 & \underline{9.33} & \underline{9.05} & \underline{9.56} & 9.24 & 8.91 & 9.31 & \underline{9.75} & \underline{9.51} & \underline{9.73} & \underline{9.29} & \underline{9.04} & \underline{9.49} \\

& Wan2.7-Image-Pro
& 9.18 & 8.99 & 9.19 & 8.78 & 8.40 & 9.26 & 8.86 & 8.85 & 8.58 & 7.31 & 7.10 & 9.14 & 7.40 & 7.24 & 8.79 & 8.53 & 8.32 & 9.29 & 8.39 & 8.17 & 9.04 & 8.66 & 8.58 & 9.15 & 8.40 & 8.21 & 9.07 \\

& \multicolumn{28}{l}{\cellcolor{gray!12}\textbf{Open-source methods}} \\
& FireRed-Image-Edit-1.1
& 1.18 & 1.09 & 1.47 & 4.78 & 4.34 & 2.27 & 6.58 & 6.56 & 2.33 & 1.86 & 1.77 & 1.54 & 1.54 & 1.44 & 1.26 & 1.81 & 1.68 & 1.64 & 1.23 & 1.21 & 1.14 & 1.00 & 1.00 & 1.00 & 2.30 & 2.20 & 1.54 \\

& Qwen-Image-Edit-2511
& 3.00 & 2.41 & 5.16 & 7.99 & 7.77 & 6.91 & 8.14 & 7.90 & 5.41 & 7.23 & 6.75 & 7.87 & 5.16 & 4.42 & 5.67 & 5.01 & 4.54 & 5.26 & 4.48 & 4.17 & 4.93 & 2.91 & 2.81 & 3.38 & 5.34 & 4.95 & 5.48 \\

& FLUX2-klein-base-9B
& 5.84 & 4.75 & 6.32 & 6.40 & 6.32 & 5.68 & 8.25 & 8.18 & 6.94 & 4.52 & 4.40 & 9.11 & 4.06 & 3.78 & 5.19 & 5.58 & 5.10 & 6.66 & 5.97 & 5.57 & 7.02 & 4.35 & 4.26 & 4.64 & 5.65 & 5.30 & 6.52 \\

\rowcolor{green!8}
& \textbf{\method{} (ours)}
& \textbf{9.88} & \textbf{9.91} & \underline{9.89} & \textbf{9.19} & \textbf{9.19} & \underline{9.65} & \textbf{9.82} & \underline{9.70} & \textbf{9.61} & \textbf{9.96} & \textbf{9.72} & \textbf{9.78} & \textbf{9.92} & \textbf{9.73} & \textbf{9.95} & \textbf{9.78} & \textbf{9.51} & \textbf{9.89} & \textbf{9.91} & \textbf{9.63} & \textbf{9.80} & \textbf{9.92} & \textbf{9.67} & \textbf{9.75} & \textbf{9.82} & \textbf{9.62} & \textbf{9.85} \\
\bottomrule
\end{tabular}%
}
\vspace{-4mm}
\end{table*}

In the precise setting, where region ambiguity is largely removed, \method further improves to $9.82/9.62/9.85$ overall in IA/EF/BP. It outperforms strong closed-source baselines including Nano Banana Pro ($9.18/9.01/9.49$) and Seedream 5.0 ($9.29/9.04/9.49$), with consistent gains across operation types. The advantage remains pronounced on source--target-sensitive edits such as \op{move} ($9.96/9.72/9.78$) and \op{exchange} ($9.92/9.73/9.95$), and performance stays high as the number of requested operations increases. These results validate the separation of fuzzy instruction understanding from grounded local execution.

\begin{figure}[t]
  \centering
  \includegraphics[width=1.0\linewidth]{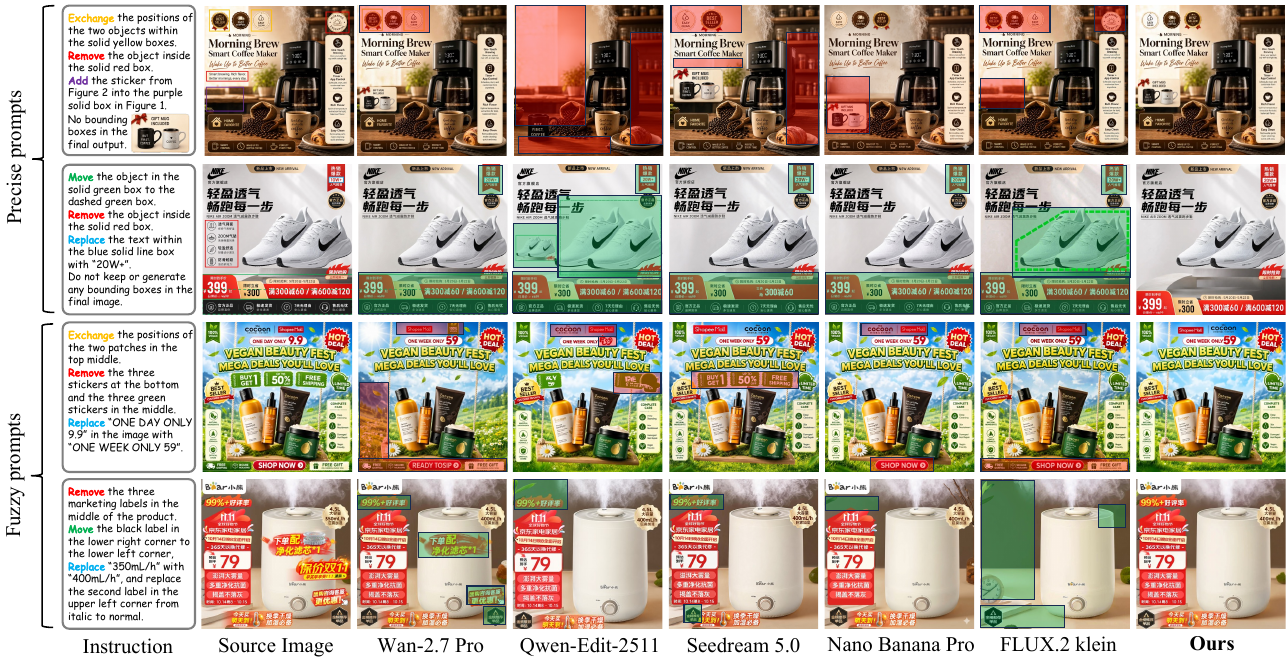}
  \vspace{-10pt}
  \caption{Qualitative comparison on image editing tasks. Highlighted regions indicate areas where the editing instruction is not fully satisfied. Our method better follows the target edit while preserving irrelevant image regions.}
  \vspace{-10pt}
  \label{fig:qualitative_results}
\end{figure}

\textbf{Qualitative Results.}
Fig.~\ref{fig:qualitative_results} shows representative fuzzy and precise editing examples. Existing editors can often generate visually plausible e-commerce posters, but they remain brittle when a request contains multiple region-specific operations. Under fuzzy prompts, competing methods frequently complete only part of the request, confuse source and target roles in \op{move} or \op{exchange}, or introduce undesired artifacts such as visible boxes. Under precise prompts, where region information is already available, several baselines still miss local removals or replacements and sometimes alter non-target product or background regions. In contrast, \method decomposes each request into a grounded edit agenda and executes the sub-operations with operation-aware masks, leading to more complete fulfillment of \op{add}, \op{remove}, \op{move}, \op{exchange}, and \op{replace} instructions while better preserving the original layout, product appearance, and irrelevant background content.

\begin{wrapfigure}{r}{0.41\textwidth}
    \vspace{-5mm}
  \begin{center}
    \includegraphics[width=0.41\textwidth]{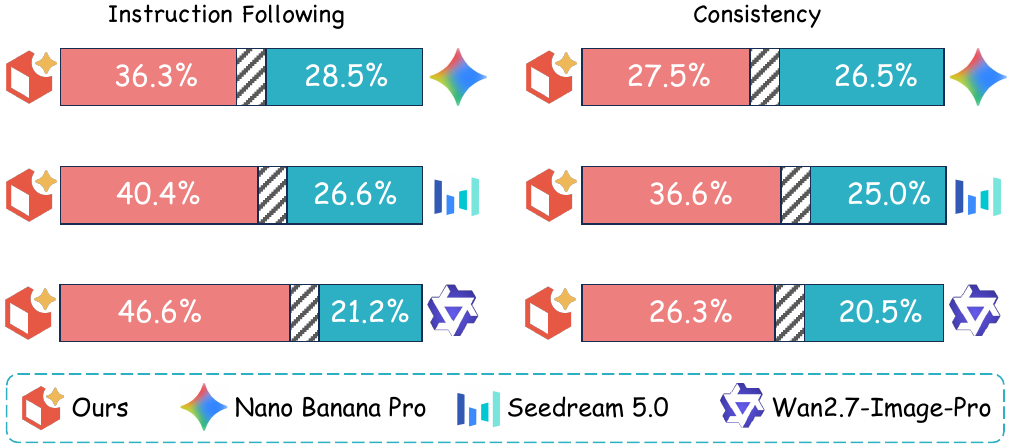}
  \end{center}
  \vspace{-6pt} 
  \caption{User study between our method and closed-source models regarding Instruction Following and Consistency.}
  \label{fig:user_study}
  \vspace{-10pt} 
\end{wrapfigure}

\textbf{User Study.} We conduct a head-to-head human evaluation on 1,000 fuzzy-instruction cases against Nano Banana Pro, Seedream 5.0, and Wan2.7-Image-Pro. As shown in Fig.~\ref{fig:user_study}, \method is consistently preferred for instruction following ($36.3\%$, $40.4\%$, and $46.6\%$ vs. $28.5\%$, $26.6\%$, and $21.2\%$) and also slightly favored for consistency ($27.5\%$, $36.6\%$, and $26.3\%$ vs. $26.5\%$, $25.0\%$, and $20.5\%$). This demonstrates that our framework can effectively adhere to multiple task instructions while maintaining high visual consistency throughout multi-task editing processes.

\subsection{Ablation Study}

Table~\ref{tab:ablation_study} studies the contribution of the staged agent--editor pipeline in \method. The upper block evaluates direct image editing without a VLM agent. The baseline image editor performs reasonably on single-operation edits but drops sharply on multi-operation cases, with only $3.783$ IA and $3.303$ EF. Fuzzy-instruction SFT substantially improves the overall scores from $5.160/4.530/6.650$ to $7.885/7.557/8.156$ in IA/EF/BP, showing that task-specific supervision is important for localized editing and background preservation. The lower block further introduces VLM-agent planning with our mask-conditioned editor. Even without agent training, the VLM agent improves compositional handling, reaching $6.017$ IA and $5.540$ EF on multi-operation cases. After SFT, the agent raises the overall score to $8.528/8.079/8.768$, confirming that learned planning and grounding are central to reliable multi-operation editing. Adding GRPO yields the best single-task IA and EF ($9.126$ and $8.814$), while joint agent--editor RL achieves the strongest multi-task and overall results, reaching $8.137/7.682/8.308$ on multi-operation cases and $8.664/8.247/8.871$ overall. This indicates that GRPO improves local decision quality, whereas joint training better aligns the planner and editor for compositional execution.

\begin{table}[t]
\centering
\caption{
Ablation study of direct editing baselines and our staged agent--editor pipeline. Best and second-best results are highlighted in \textbf{bold} and \underline{underlined}.
}
\label{tab:ablation_study}
\setlength{\tabcolsep}{4pt}
\renewcommand{\arraystretch}{1.08}
\resizebox{\textwidth}{!}{%
\begin{tabular}{l|c| c|c|c|c|c|c|c|c|c}
\toprule
\multirow[c]{2}{*}{\textbf{Method}}
& \multirow[c]{2}{*}{\textbf{Training Stage}}
& \multicolumn{3}{c|}{\textbf{Single-task}}
& \multicolumn{3}{c|}{\textbf{Multi-task}}
& \multicolumn{3}{c}{\textbf{Overall}} \\
\cmidrule(lr){3-5}
\cmidrule(lr){6-8}
\cmidrule(lr){9-11}
& & IA & EF & BP
& IA & EF & BP
& IA & EF & BP \\
\midrule

\rowcolor{gray!12}
\multicolumn{11}{l}{\emph{w/o VLM agent}} \\
Baseline image editor (w/o training)
& -
& 6.286 & 5.538 & 7.866
& 3.783 & 3.303 & 5.183
& 5.160 & 4.530 & 6.650 \\
Baseline image editor + fuzzy-instruction SFT
& -
& 8.540 & 8.260 & 8.884
& 7.070 & 6.728 & 7.383
& 7.885 & 7.557 & 8.156 \\

\midrule
\rowcolor{gray!12}
\multicolumn{11}{l}{\emph{w/ VLM agent and mask-conditioned editor}} \\
VLM agent (w/o training)
& Stage 2
& 7.946 & 7.612 & 8.855
& 6.017 & 5.540 & 7.488
& 6.883 & 6.463 & 8.082 \\
VLM agent SFT 
& Stages 1--2
& 9.109 & 8.734 & \textbf{9.387}
& 7.884 & 7.397 & 8.146
& 8.528 & 8.079 & 8.768 \\
VLM agent SFT + GRPO 
& Stages 1--2
& \textbf{9.126} & \textbf{8.814} & 9.338
& \underline{7.973} & \underline{7.482} & \underline{8.195}
& \underline{8.618} & \underline{8.200} & \underline{8.801} \\
VLM agent SFT + GRPO + joint agent--editor RL
& Stages 1--3
& \underline{9.112} & \underline{8.764} & \underline{9.382}
& \textbf{8.137} & \textbf{7.682} & \textbf{8.308}
& \textbf{8.664} & \textbf{8.247} & \textbf{8.871} \\
\bottomrule
\end{tabular}%
}
\vspace{-3mm}
\end{table}

\begin{wraptable}{r}{0.42\textwidth}
  \centering
  \vspace{-5mm}
  \captionsetup{font=small}
  \caption{Ablation study on multi-turn editing.}
  \label{tab:ablation_study_multi_turn}
  \setlength{\tabcolsep}{10pt}
  \resizebox{\linewidth}{!}{%
    \begin{tabular}{l|ccc}
    \toprule
    \textbf{Model} & \textbf{IA} & \textbf{EF} & \textbf{BP} \\
    \midrule
    Stage1-one-turn     & 6.89 & 5.98 & 7.86 \\
    Stage3-one-turn     & 7.04 & 6.23 & \underline{8.14} \\
    Stage3-fixed-turn   & \underline{7.21} & \underline{6.37} & 8.08 \\
    \rowcolor{gray!15}
    Stage3-reflection   & \textbf{7.29} & \textbf{6.47} & \textbf{8.42} \\
    \bottomrule
    \end{tabular}%
  }
  \vspace{-3mm}
\end{wraptable}

To evaluate the reflection-driven multi-turn mechanism, we compare different strategies in Table~\ref{tab:ablation_study_multi_turn}. Relative to the Stage 1 one-turn baseline, Stage 3 one-turn inference improves IA and EF from $6.89/5.98$ to $7.04/6.23$, highlighting the benefit of the jointly optimized agent-editor policy. Using a fixed-turn strategy further increases IA/EF to $7.21/6.37$, showing that additional execution rounds help recover incomplete operations. Our full reflection-based policy achieves the highest scores across all metrics, with IA, EF, and BP, confirming that adaptive state diagnosis, which allows the agent to decide whether to stop, continue, or roll back, is more robust than using a fixed number of refinement rounds. More details are presented in Appendix~\ref{sec:multi-turn}.

\vspace{-1mm}
\section{Conclusion}
\vspace{-1mm}

In this paper, we have presented \method, a novel agentic framework that redefines e-commerce image editing from a single-step generation task into a systematic, multi-operation task completion process. By decoupling cognitive reasoning from generative execution, our framework leverages a VLM-based agent to translate fuzzy user instructions into structured agendas and manages the editing workflow through a self-reflective loop. This design effectively addresses the brittleness of existing one-shot editors when faced with complex, compositional requests typical of real-world e-commerce scenarios.
Furthermore, we introduced a unified data pipeline and the \bench benchmark to facilitate the training and evaluation of grounded, multi-turn editing. Our experimental results demonstrate that \method consistently outperforms both state-of-the-art open-source and proprietary models, particularly in terms of instruction adherence, regional fidelity, and long-term reliability through its rollback-and-retry mechanism.

\clearpage

\bibliographystyle{unsrtnat}
\bibliography{ref}

\newpage
\appendix
\clearpage

\section*{\Large{Appendix}}

In this work, we propose GMO-E\textsuperscript{2}DIT, an agentic framework for grounded multi-operation e-commerce image editing. Due to space constraints in the main manuscript, several technical details, data construction processes, and additional qualitative results were omitted. The purpose of this supplementary document is to provide these comprehensive details to facilitate a deeper understanding and reproducibility of our work. Specifically, this material provides the following contents:

\begin{itemize}
\item \textbf{Method Details (Section~\ref{sec:supp_method_details}):} Detailed explanations and mathematical formulations of the reward designs for both the initial planning and multi-turn joint reinforcement learning stages;
\item \textbf{Data Pipeline Details (Section 2):} Comprehensive technical details of the multi-stage perception and generation pipeline utilized for editable element mining and structural synthesis;
\item \textbf{Benchmark Details (Section~\ref{app:benchmark_details}):} Elaboration on the task settings, precise and fuzzy evaluation protocols, and VLM-based scoring criteria of the constructed benchmark;
\item \textbf{Experimental Details (Section~\ref{implementation_details}):} Additional training hyperparameters, optimization strategies, and baseline configurations for both the VLM agent and the mask-conditioned editor;
\item \textbf{More Experimental Results (Section E):} Extended qualitative comparisons demonstrating the advantages of multi-turn instruction execution over single-turn baselines;
\item \textbf{Limitations (Section F):} A discussion of the current boundaries and limitations of our proposed framework.
\end{itemize}

\section{Method Details}
\label{sec:supp_method_details}

\subsection{Stage 1 Reward Design: Planning and Reflection Rewards}

During Stage 1, the agent acquires fundamental planning and diagnostic capabilities through supervised fine-tuning (SFT) and initial reinforcement learning. The reward signals during this phase are applied at a single-turn level.

\textbf{Planning Rewards ($R_{\mathrm{plan}}$)} comprise four verifiable metrics designed to evaluate the generated edit agenda:

\begin{equation}
    R_{\mathrm{plan}} = w_1 R_{\mathrm{JSON}} + w_2 R_{\mathrm{order}} + w_3 R_{\mathrm{type}} + w_4 R_{\mathrm{bbox}}
\end{equation}

where:
\begin{itemize}
    \item $R_{\mathrm{JSON}}$ verifies that the generated edit agenda is strictly parsable and accurately populates all requisite structural fields.
    
    \item $R_{\mathrm{order}}$ evaluates the logical sequencing of operations, heavily penalizing illogical temporal sequences 
    (e.g., executing an \op{add} operation before a prerequisite \op{remove} operation).
    
    \item $R_{\mathrm{type}}$ assesses the accuracy of the predicted operation types 
    (e.g., \op{move}, \op{add}, \op{remove}) against the ground truth.
    
    \item $R_{\mathrm{bbox}}$ measures the spatial alignment of the predicted source and target bounding boxes. 
    This is quantified using the Intersection over Union (IoU) between the predicted bounding box $B_p$ and the ground-truth box $B_{gt}$:
\end{itemize}

\begin{equation}
    \text{IoU}(B_p, B_{gt}) = \frac{\text{Area}(B_p \cap B_{gt})}{\text{Area}(B_p \cup B_{gt})}
\end{equation}

The coefficients $w_1$, $w_2$, $w_3$, and $w_4$ are hyperparameters that balance the contribution of each reward component. The default weights are set as $w_1 = 0.3$, $w_2 = 0.2$, $w_3 = 0.2$, and $w_4 = 0.3$.

\textbf{Reflection Rewards ($R_{\mathrm{reflect}}$)} evaluate the agent's diagnostic phase using three task-oriented signals:

\begin{equation}
R_{\mathrm{reflect}} =  w_5 R_{\mathrm{format}} + w_6 R_{\mathrm{status}} + w_7 R_{\mathrm{next}} 
\end{equation}

where:
\begin{itemize}
    \item $R_{\mathrm{format}}$ ensures that the structured diagnostic reflection strictly adheres to the required JSON schema.
    
    \item $R_{\mathrm{status}}$ evaluates the accuracy of the predicted intermediate state classification 
    (\op{success}, \op{continue}, or \op{rollback}).
    
    \item $R_{\mathrm{next}}$ assesses whether the agent correctly identifies the remaining task IDs when the current status is not \op{success}.
\end{itemize}

The coefficients $w_5$, $w_6$, and $w_7$ are hyperparameters that balance the contribution of each reward component. The default weights are set as $ w_5 = 0.4$, $w_6 = 0.3$, and $w_7 = 0.3$.

Additionally, we incorporate Chain-of-Thought (CoT) reasoning during the reflection phase to explicitly guide the agent through intermediate logical deliberation before outputting the final state diagnosis.

\subsection{Stage 3 Reward Design: Multi-turn Joint Reinforcement Training Rewards}

Stage 3 is the primary multi-turn training stage, where the agent interacts with the frozen mask-conditioned editor in a multi-turn loop. The rewards in this stage are dynamic and trajectory-based, focusing on the convergence of the entire editing session. In this multi-turn setting, the reward mechanism is bipartite, consisting of an initial planning reward and subsequent reflection rewards guided by a stronger VLM.

\textbf{Planning Reward.} During the initial turn of the editing trajectory, the agent generates the structured edit agenda. To maintain the foundational planning capabilities established earlier, the planning reward $R_{\mathrm{plan}}$ remains strictly consistent with the formulation in Stage 1. This ensures that the generated agenda maintains rigorous structural, sequential, and spatial alignment with the ground truth before execution begins.

\textbf{VLM-Guided Reflection Rewards.} For subsequent turns, we employ a stronger, more capable VLM as a Reward Model (RM) to evaluate the intermediate edited images. The RM independently assesses the current editing status after each editor execution. The reflection reward is then dynamically calculated based on the consistency between the agent's self-diagnosis and the RM's objective evaluation.

Specifically, the status classification reward $R_{\mathrm{status}}$ prioritizes reliable convergence over greedy completion through the following criteria:
\begin{itemize}
    \item \textbf{Success Reward (High Positive):} Awarded only when both the agent and the RM agree that all tasks are completed correctly.
    \item \textbf{Continue/Progress Reward (Cumulative Positive): }To encourage safe, incremental edits, the agent receives positive reinforcement for correctly identifying a \op{continue} state that aligns with the RM's assessment. This allows the system to build on partial successes rather than failing a complex request in one go.
    \item \textbf{Rollback Penalty (Heavy Negative):} If the agent triggers a \op{rollback}, it incurs a "rollback tax"—a penalty for inefficient execution. However, a "correct rollback" (where the agent identifies a fatal error corroborated by the RM and chooses to restart) is penalized far less severely than a "false success," thereby prioritizing product integrity.
    \item \textbf{Early Termination Penalty (Extreme Negative):} We strictly penalize the agent if it erroneously predicts \op{success} when the RM detects missed operations or visual corruption. This mechanism is crucial to prevent the model from exploiting a shortcut by indiscriminately predicting success to terminate the loop early.
\end{itemize}

Beyond these status-based rewards, the reflection phase strictly retains the format reward ($R_{\mathrm{format}}$) and the next-task identification reward ($R_{\mathrm{next}}$) introduced in Stage 1. This ensures that the structured diagnosis remains parsable and logically consistent for directing the subsequent turns.

\textbf{Trajectory-Level Advantage Allocation.} Unlike standard RL, the multi-turn reward is assigned to specific token spans across different turns. For a trajectory with $T$ rounds, the total reward is a combination of the initial planning quality (Turn 1) and the subsequent VLM-guided reflection accuracy (Turns 2 to $T$). We utilize a step-level advantage calculation to ensure that the reward from Round $t$ specifically reinforces the policy parameters responsible for that exact turn's output, preventing negative reflection rewards from degrading competent planning behaviors.

\section{Details of Dataset Pipeline}

To construct a high-quality, fine-grained dataset for e-commerce image editing, our editable element mining pipeline leverages a suite of state-of-the-art vision and language models. This multi-stage perception and generation pipeline is designed to accurately decompose complex e-commerce posters into independent, structured semantic assets. The specific models and their corresponding roles are detailed below:

\textbf{Product and Graphic Object Localization: }We employ the YOLOv10~\cite{wang2024yolov10} object detector to identify and localize primary and salient visual elements within the source images. This model is responsible for generating precise bounding boxes for product instances, brand logos, decorative stickers, and other foreground graphic objects, serving as the foundational spatial prior for subsequent extraction.
    
\textbf{Layer Segmentation and Matting: }To ensure the extracted visual assets are of high fidelity and seamlessly reusable, we utilize a segmenter based on SAM2~\cite{ravi2024sam}. Conditioned on the bounding boxes provided by the localization stage, SAM2 performs pixel-level semantic segmentation and high-quality alpha matting. This process isolates the objects from complex backgrounds, yielding sharp, transparent masks that are crucial for downstream operations like \textit{move} and \textit{exchange}.

\textbf{Text and Layout Detection:} Recognizing the critical importance of text in e-commerce images, we rely on PP-OCRv5~\cite{cui2026pp} for comprehensive text analysis. This robust OCR model extracts both the textual content and the geometric layout of all text blocks. Capturing this text-specific metadata is essential for enabling localized, attribute-level modifications.

\textbf{Background Recovery: }Once the foreground elements (products, graphics, and text) are extracted, the remaining image contains missing regions. To create a clean, element-free background canvas, we deploy Flux2-klein-9B~\cite{blackforestlabs2025flux2}, a powerful diffusion-based inpainting model. This model intelligently fills the voids by referencing the surrounding context, resulting in a natural, semantically consistent background that serves as the foundation for generative recomposition.
    
\textbf{Structural Synthesis:} Finally, we utilize Qwen3.5~\cite{team2026qwen3}, a highly capable Multimodal Large Language Model, as the central integration engine. Acting as the cognitive orchestrator, Qwen3.5 synthesizes the parsed visual attributes, extracted textual content, and layout coordinates from the preceding modules. It translates these discrete, multi-modal components into a structured, machine-interpretable format (e.g., HTML). This hierarchical representation captures the underlying semantic logic and spatial constraints of the original poster, effectively forming the grounded "edit program" used to guide the training of both the VLM agent and the mask-conditioned editor.

\section{Details of Benchmark}
\label{app:benchmark_details}

\subsection{Benchmark Construction}
\label{app:benchmark_construction}

Following recent benchmark designs that report both task-wise and difficulty-wise results~\cite{chen2026macro,qu2025replan}, we construct \bench as a held-out benchmark for grounded multi-operation e-commerce image editing. The benchmark contains 1,000 examples covering five atomic operations: \op{add}, \op{remove}, \op{move}, \op{exchange}, and \op{replace}. It includes 500 single-operation cases, with 100 examples for each atomic operation, and 500 multi-operation cases, including 200 two-operation, 200 three-operation, and 100 four-operation requests. This split allows us to separately analyze primitive operation execution and compositional generalization.

Each benchmark example contains a source product image, a target image for reference-enabled diagnostics, an optional patch image for insertion tasks, an operation-type string, and structured region annotations. For fuzzy evaluation, we additionally provide a natural-language user instruction for each sample. The benchmark is disjoint from the training data at the product and edit-program level.

\begin{table}[t]
\centering
\caption{Composition of \bench. Single-operation cases are balanced across five atomic operations, while multi-operation cases evaluate compositional editing under multiple local constraints.}
\label{tab:benchmark_composition_appendix}
\setlength{\tabcolsep}{7pt}
\renewcommand{\arraystretch}{1.08}
\begin{tabular}{lcc}
\toprule
\textbf{Group} & \textbf{Subgroup} & \textbf{Number of samples} \\
\midrule
Single-operation & \op{add} / \op{remove} / \op{move} / \op{exchange} / \op{replace} & $5\times100=500$ \\
\midrule
\multirow{3}{*}{Multi-operation} & Two operations & 200 \\
& Three operations & 200 \\
& Four operations & 100 \\
\midrule
Total & -- & 1,000 \\
\bottomrule
\end{tabular}
\end{table}

\subsection{Operation Annotation Format}
\label{app:operation_annotation}

The precise protocol uses a structured edit program converted from region annotations. Each operation is represented by its type, semantic label(s), and the required source and/or target bounding boxes in \texttt{xywh} format. The five operation types are defined as follows:

\begin{table}[t]
\centering
\caption{Structured annotation fields used by the precise evaluation protocol.}
\label{tab:operation_format_appendix}
\setlength{\tabcolsep}{5pt}
\renewcommand{\arraystretch}{1.08}
\resizebox{\linewidth}{!}{%
\begin{tabular}{lll}
\toprule
\textbf{Operation} & \textbf{Required region information} & \textbf{Editing requirement} \\
\midrule
\op{add} & Target box and inserted-object label; optional patch image & Insert the specified patch or object at the target region. \\
\op{remove} & Source box and object label & Remove the specified object while preserving surrounding content. \\
\op{move} & Source box, target box, and object label & Move the object from the source region to the target region. \\
\op{exchange} & Source/target boxes for two labeled objects & Swap the two specified objects or regions. \\
\op{replace} & Source/target box, label, and replacement constraints & Replace local content, attributes, text, color, or layout as specified. \\
\bottomrule
\end{tabular}%
}
\end{table}

For compositional examples, the edit program contains multiple operation entries. This makes it possible to evaluate whether a model misses sub-operations, edits the wrong region, confuses source and target roles, or introduces spillover outside the intended local regions.

\subsection{Precise and Fuzzy Evaluation Protocols}
\label{app:precise_fuzzy_protocols}

We evaluate all models under two complementary protocols.

\textbf{Fuzzy protocol.} The model receives the source image and a natural-language request written in a user-like form. The instruction may contain approximate spatial descriptions, referring expressions, or implicit operation bindings. For \op{add} cases, an insertion patch can be provided when the task requires copying a logo, text, graphic, or product element. This protocol evaluates the full editing pipeline, including instruction understanding, operation decomposition, target localization, and local editing.

\textbf{Precise protocol.} The model receives the source image and a structured region-level edit description derived from the annotations in Table~\ref{tab:operation_format_appendix}. For our method, this program is further compiled into operation-aware masks and optional patch references. This protocol removes most grounding ambiguity and isolates the editor's ability to execute local operations faithfully.

\subsection{VLM-as-Judge Evaluation}
\label{app:vlm_judge}

Inspired by recent automatic evaluation protocols for generation and editing benchmarks~\cite{chen2026macro,qu2025replan}, we use a fixed VLM-as-Judge protocol for all methods. The judge receives images in a fixed order: the source image, the edited result, an optional ground-truth target image in the reference-enabled setting, and an optional patch image for \op{add} tasks. Unless otherwise specified, the reported results use the same Gemini-3 Pro judge as the main paper; the implementation keeps the judge temperature at $0$ and uses the same prompt and parsing rules for every model.

The judge outputs three integer scores in $[1,10]$: \textbf{Instruction Accuracy (IA)}, \textbf{Edit Fidelity (EF)}, and \textbf{Background Preservation (BP)}. IA measures whether all requested operations are applied to the correct objects and locations. EF measures the semantic correctness and visual quality of the edited regions, including patch faithfulness for insertion tasks. BP measures whether non-target regions remain unchanged.

\begin{table}[t]
\centering
\caption{Score bands used by the VLM evaluator.}
\label{tab:vlm_score_bands_appendix}
\setlength{\tabcolsep}{4pt}
\renewcommand{\arraystretch}{1.08}
\resizebox{\linewidth}{!}{%
\begin{tabular}{lcccc}
\toprule
\textbf{Metric} & \textbf{10} & \textbf{7--9} & \textbf{4--6} & \textbf{1--3} \\
\midrule
IA & All operations exactly executed & Minor omission or placement error & Partial execution or some wrong operations & Instruction largely not executed \\
EF & Correct and natural edited content & Minor visual/style/text issues & Recognizable but noticeably imperfect & Wrong, distorted, or missing edit content \\
BP & Pixel-level background preservation & Imperceptible drift & Visible unintended change but structure kept & Major unintended background alteration \\
\bottomrule
\end{tabular}%
}
\end{table}

\subsection{Evaluation Prompt Template}
\label{app:evaluation_prompt}

We use one unified prompt template for both protocols. The placeholders are filled according to the instruction mode, whether a ground-truth target is used, and whether the sample contains an insertion patch.

\begin{quote}
\small
\textbf{Role.} You are an expert image editing evaluator. You will be shown multiple images in this exact order: source image, edited image, optional ground-truth target image, and optional patch image.\\[2pt]
\textbf{Task.} Evaluate the quality of the edited image against the benchmark instruction, the source image, and optional references.\\[2pt]
\textbf{Benchmark Instruction.} \texttt{\{benchmark\_instruction\}}\\[2pt]
\textbf{Evaluation Dimensions.}\\
1. \texttt{instruction\_accuracy}: Did the model execute the specified edit operations?\\
2. \texttt{edit\_fidelity}: How well does the edited result satisfy the requested edit inside the target regions? For insertion tasks, also judge whether the inserted content matches the patch image.\\
3. \texttt{background\_preservation}: Outside the requested edit regions, is the edited image unchanged compared with the source image?\\[2pt]
\textbf{Instructions.} Read the benchmark instruction carefully. In the precise protocol, examine each bounding box and reference region indices in the reasoning. In the fuzzy protocol, identify the intended edit from the natural-language request and visible evidence. Be objective and do not reward changes that were not requested.\\[2pt]
\textbf{Response Format.} Return exactly three integer scores in $[1,10]$ followed by a short explanation:

\verb|<scores>|\\
\verb|instruction_accuracy: [int 1--10]|\\
\verb|edit_fidelity: [int 1--10]|\\
\verb|background_preservation: [int 1--10]|\\
\verb|</scores>|\\
\verb|<reasoning>|\\
\verb|2--4 sentences referencing the instruction and visible evidence.|\\
\verb|</reasoning>|
\end{quote}

For the precise protocol, \texttt{\{benchmark\_instruction\}} is a formatted list of operation entries and bounding boxes. For the fuzzy protocol, it is the natural-language user request. The same response parser is used in both settings, and a sample is treated as invalid if any of the three scores cannot be parsed as an integer in $[1,10]$.

\subsection{Aggregation and Reporting}
\label{app:aggregation}

After scoring, we average IA, EF, and BP over all valid samples. We also report fine-grained results by single-operation type and by the number of operations in compositional cases. Specifically, single-operation samples are grouped by their atomic operation type, while multi-operation samples are grouped into two-, three-, and four-operation cases. This aggregation exposes complementary failure modes: low IA indicates grounding or operation-binding failures, low EF indicates poor local rendering or semantic mismatch, and low BP indicates unintended changes to non-target regions.

\section{Experiments Details}
\label{implementation_details}

Our image editing backbone is initialized from Flux2-Klein-9B~\cite{blackforestlabs2025flux2}, while the VLM editing agent is based on Qwen3.5-9B~\cite{team2026qwen3}. The VLM handles instruction parsing, region-level planning, and visual reflection, whereas the editor receives the source image along with operation-aware masks and optional patch references. We optimize both models using AdamW. In Stage 1, the VLM undergoes supervised fine-tuning (SFT) with LoRA of rank 32 and a learning rate of $1\times10^{-4}$ for 20K steps with a batch size of 8. The Stage 1 GRPO training uses a group size of 8, learning rate $5\times10^{-6}$, batch size 16, and runs for 1,200 steps. The image editor is fine-tuned separately with LoRA of rank 128, learning rate $1\times10^{-4}$, for 25K steps. Stage 3 employs multi-turn GRPO with a sampling group of 8, LoRA rank 16, batch size 8, learning rate $1\times10^{-5}$, and 500 training steps. Qwen3-VL-32B~\cite{bai2025qwen3} serves as the reward model, and the maximum number of execution rounds during sampling is set to 4. Unless otherwise specified, all experiments are conducted on the same constructed training dataset and evaluated on the held-out \bench split. All experiments are conducted on 16 NVIDIA B200 GPUs.

\section{More Experiment Results}
\subsection{Multi-turn Editing}
\label{sec:multi-turn}

To evaluate our multi-turn joint RL and reflection-driven mechanism, we conduct experiments on a test set containing 7–10 task editing scenarios per image, designed to validate the model’s ability to handle multi-operation, region-specific instructions. Table~\ref{tab:ablation_study_multi_turn} compares four inference configurations:
\begin{itemize}
    \item \textbf{Stage1-one-turn:} The baseline agent (after Stage 1 SFT and rule-based RL) executing the entire edit agenda in a single forward pass.
    \item \textbf{Stage3-one-turn:} The Stage 3 agent restricted to a single turn, isolating the intrinsic performance gains of joint RL training without using multi-round rollouts.
    \item \textbf{Stage3-fixed-turn:} The Stage 3 agent executing a fixed maximum number of rounds (e.g., 3), dividing tasks blindly without adaptive state reflection.
    \item \textbf{Stage3-reflection (Ours):} Our full framework. The agent autonomously evaluates intermediate states to dynamically choose to \op{success} (terminate), \op{continue} (execute remaining), or \op{rollback} (discard and retry).
\end{itemize}

\textbf{Quantitative Analysis.} As shown in Table~\ref{tab:ablation_study_multi_turn}, \textbf{Stage3-one-turn} outperforms \textbf{Stage1-one-turn} (IA: 7.04 vs. 6.89; EF: 6.23 vs. 5.98), demonstrating that joint RL inherently improves spatial planning even for single-pass generation. Extending to multiple rounds, \textbf{Stage3-fixed-turn} further elevates IA and EF (7.21 and 6.37) by successfully recovering missed operations. However, forcing fixed iterative edits drops Background Preservation (BP) from 8.14 to 8.08 due to noise accumulation from unnecessary forward passes. Our \textbf{Stage3-reflection} policy optimally resolves this trade-off. By adaptively halting completed tasks to prevent over-editing and rolling back corrupted steps, it achieves the highest scores across all metrics (7.29 IA, 6.47 EF, 8.42 BP), ensuring strict product preservation.

\begin{figure}[htbp]
  \centering
  \includegraphics[width=1.0\linewidth]{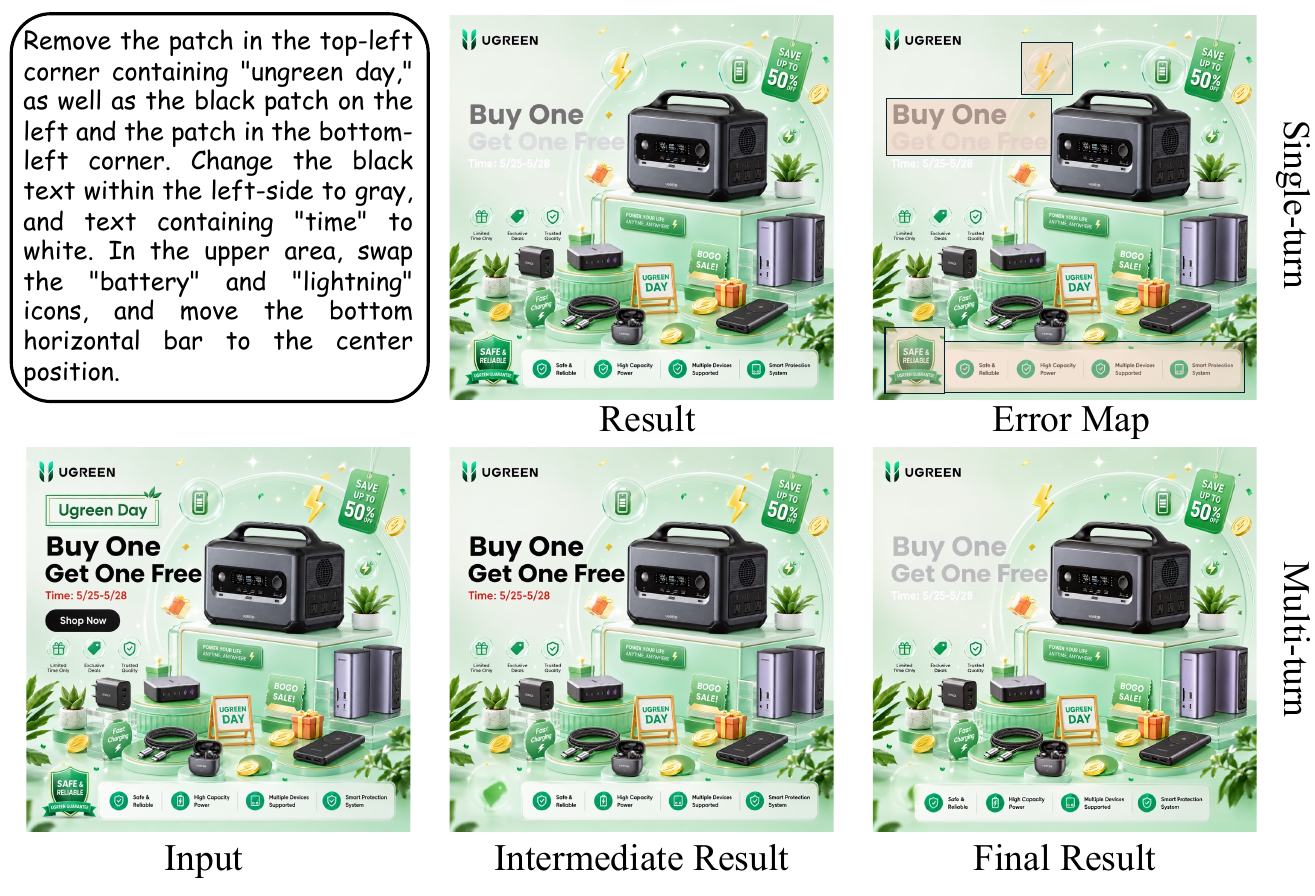} 
  \caption{Comparison of single-turn vs. multi-turn instruction execution. (Example 1)}
  \label{fig:data_pipeline_overview1}
\end{figure}

\begin{figure}[htbp]
  \centering
  \includegraphics[width=1.0\linewidth]{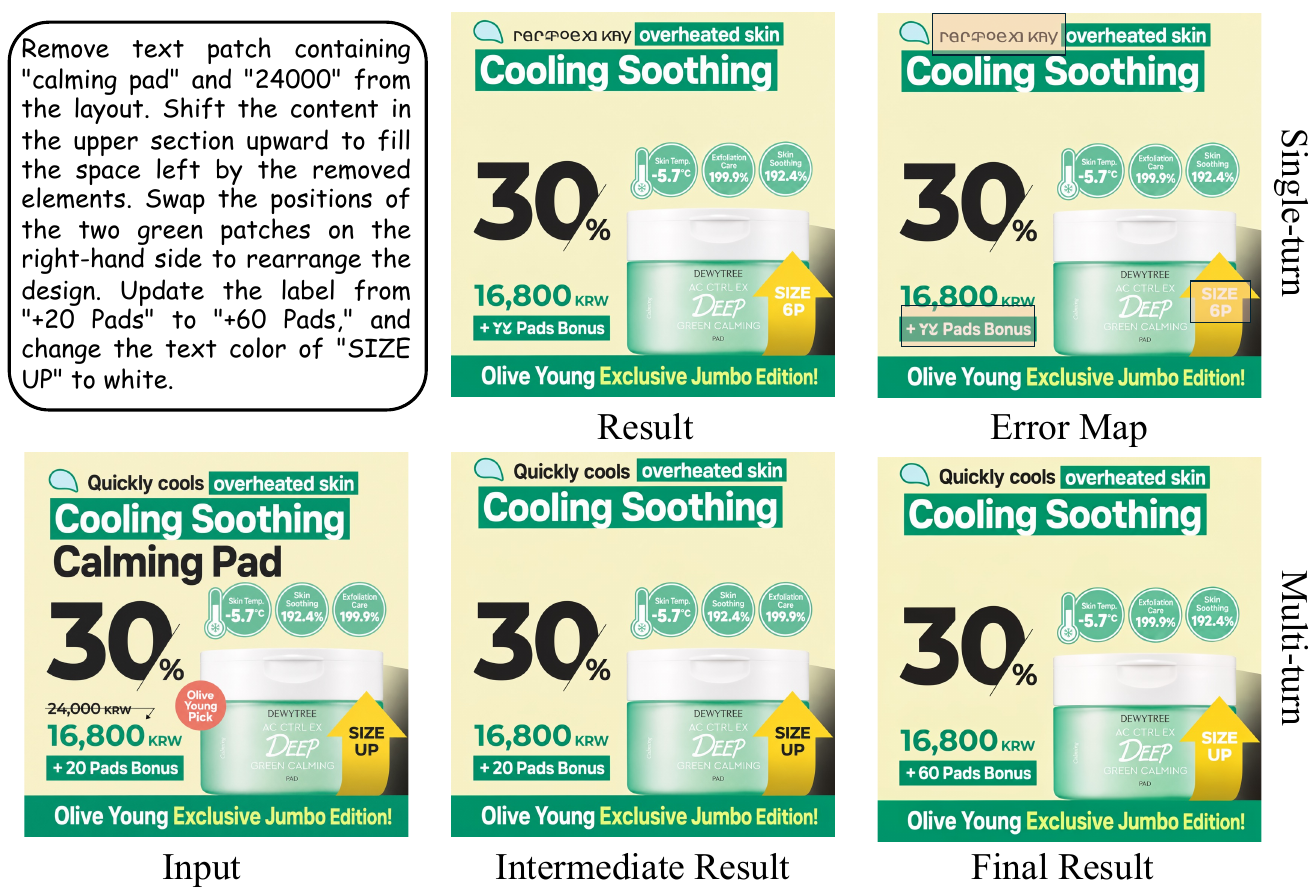} 
  \caption{Comparison of single-turn vs. multi-turn instruction execution. (Example 2)}
  \label{fig:data_pipeline_overview2}
\end{figure}

\begin{figure}[htbp]
  \centering
  \includegraphics[width=1.0\linewidth]{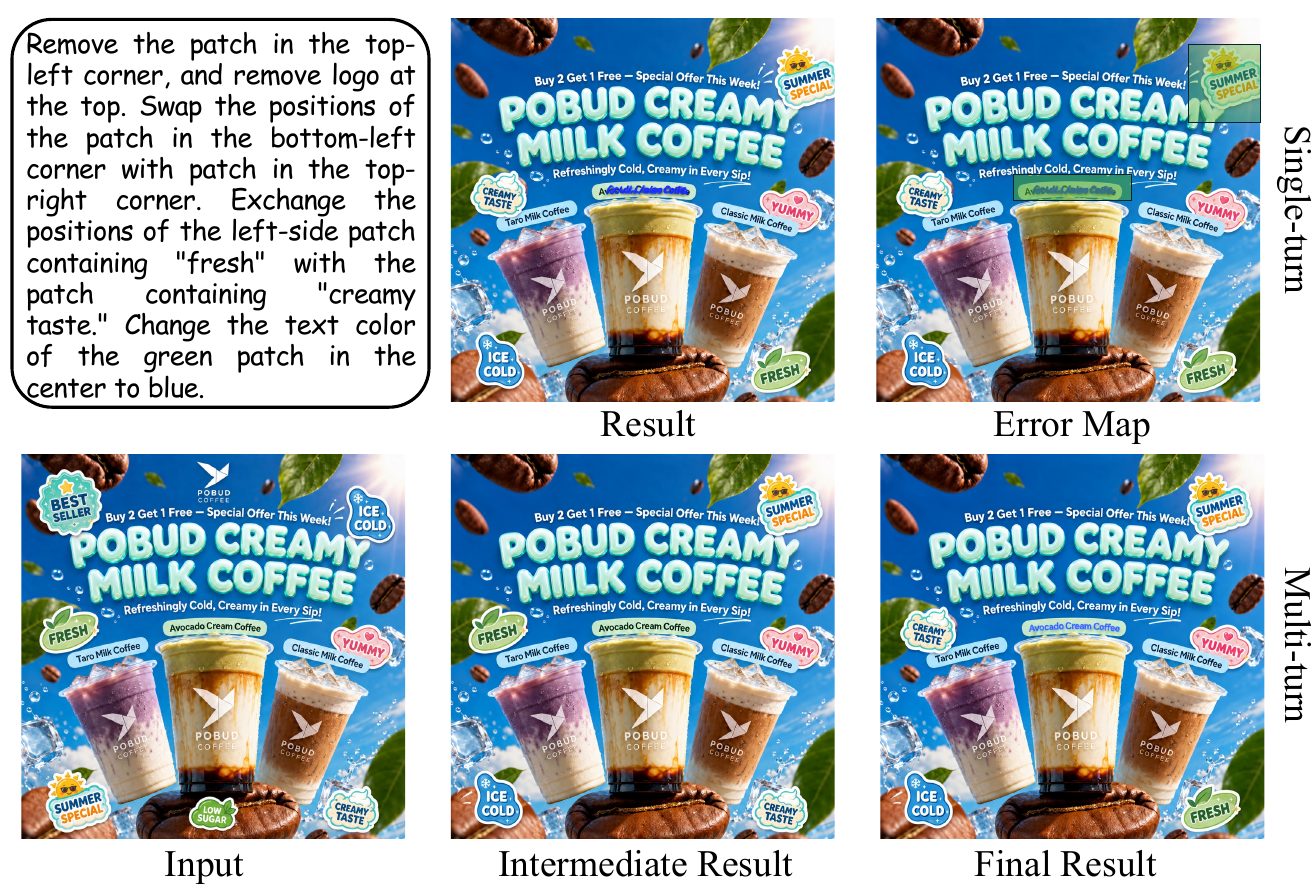} 
  \caption{Comparison of single-turn vs. multi-turn instruction execution. (Example 3)}
  \label{fig:data_pipeline_overview3}
\end{figure}

\begin{figure}[htbp]
  \centering
  \includegraphics[width=1.0\linewidth]{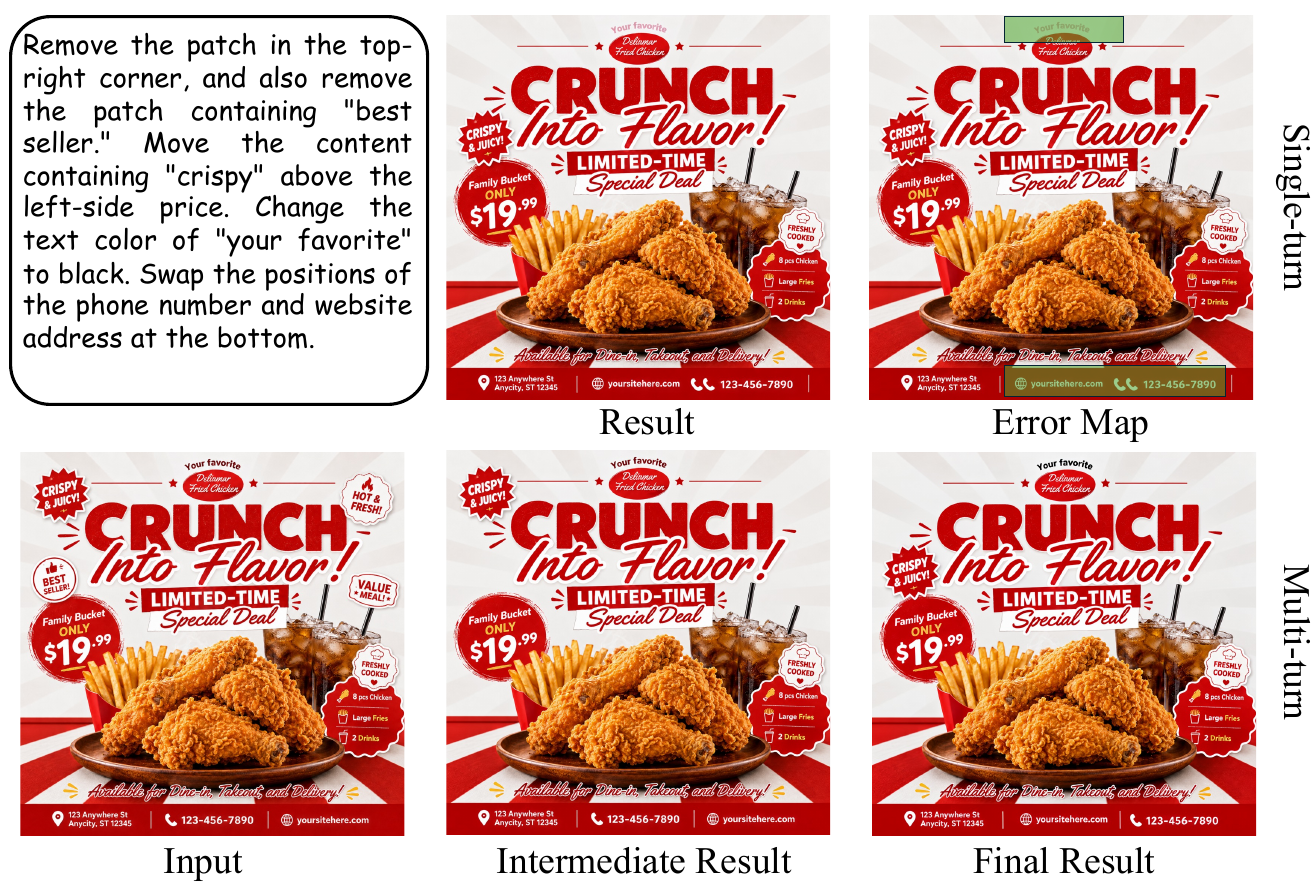} 
  \caption{Comparison of single-turn vs. multi-turn instruction execution. (Example 4)}
  \label{fig:data_pipeline_overview4}
\end{figure}

\textbf{Qualitative Analysis.} As illustrated in Fig.~\ref{fig:data_pipeline_overview1}, ~\ref{fig:data_pipeline_overview2}, ~\ref{fig:data_pipeline_overview3}, and ~\ref{fig:data_pipeline_overview4}, single-turn editors often suffer from \textit{instruction interference} on complex requests, leading to attribute bleeding, or omitted edits. Our multi-turn paradigm mitigates this by chunking requests into digestible sub-programs, highlighting two key behaviors:
1) Incremental Safety: Isolating distinct operations across turns prevents color or text styles from bleeding between adjacent regions.
2) Autonomous Recovery: If an intermediate edit distorts text or misplaces patches, the agent identifies the anomaly, triggers a \op{rollback}, and attempts an alternative sub-plan. These results validate our framework's capacity for reliable sequential planning in real-world e-commerce applications.

\subsection{Additional Qualitative Results}
\begin{figure}[htbp]
  \centering
  \includegraphics[width=1.0\linewidth]{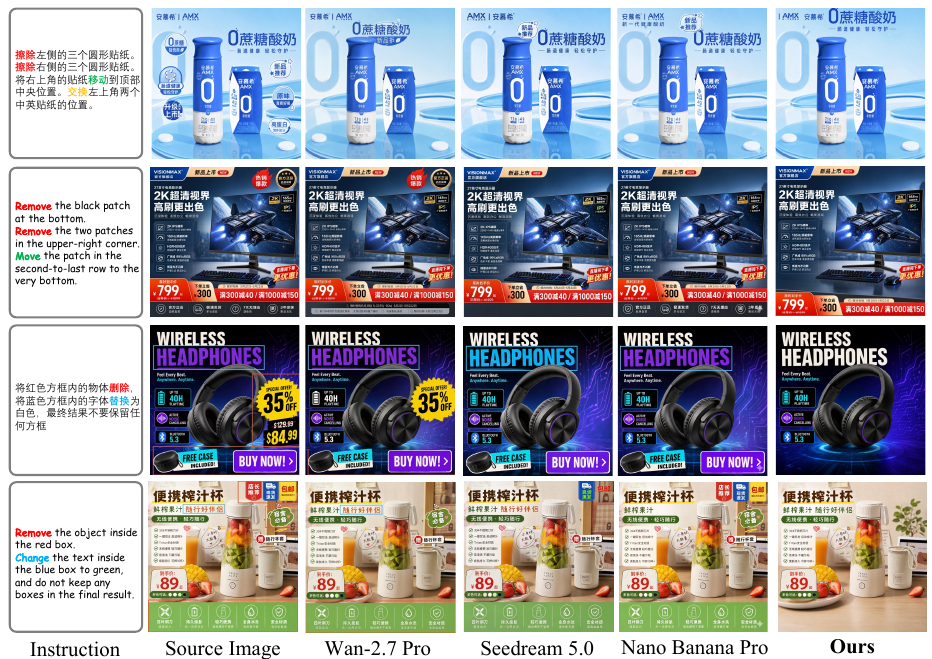} 
  \caption{More visual comparison with closed-source methods across multiple tasks.}
  \label{fig:extra_result}
\end{figure}

To further demonstrate the effectiveness of \method, we provide additional visual comparisons against strong closed-source models. Fig.~\ref{fig:extra_result} presents results across multiple multi-operation editing tasks, highlighting our framework's ability to accurately perform localized edits and preserve unmodified content.

\section{Limitations}

While our proposed \method achieves strong performance in multi-task e-commerce image editing, several limitations remain to be addressed in future work.

\textbf{Dependence on VLM Capabilities.} First, the framework’s effectiveness relies heavily on the cognitive reasoning and reflection abilities of the VLM agent. Errors in understanding complex instructions during the planning phase or misjudgments of intermediate states during reflection can compromise the entire multi-operation workflow. In practical applications, user prompts may be extremely vague, ambiguous, or linguistically unusual, leading to incorrect operation bindings, hallucinated edits, or omitted sub-tasks. These challenges highlight the need for more robust intent-resolution mechanisms.

\textbf{Dataset Limitations and Generalization.} Second, the model’s generalization is inherently constrained by the constructed dataset. Although the training pipeline and \bench benchmark cover a comprehensive set of operations (e.g., \op{add}, \op{remove}, \op{move}, \op{exchange}, \op{replace}), they are primarily derived from synthetic or semi-synthetic e-commerce posters. Real-world imagery often presents more complex layouts, diverse typography, and severe visual clutter.

\textbf{Computational Cost and Efficiency.} Finally, the multi-turn editing mechanism introduces a trade-off between reliability and computational efficiency. The reflection-driven iterative loop, which employs \op{continue} and \op{rollback} actions, substantially improves task completion and reduces error propagation, but it requires multiple sequential forward passes through both the VLM agent and the diffusion-based image editor. Consequently, the iterative process increases inference latency and computational overhead. Deploying the framework for real-time interactive editing or large-scale batch processing may therefore face practical efficiency limitations. Future work will explore model distillation and more efficient sampling strategies to mitigate these constraints.

\end{document}


\maketitle

\begin{abstract}
  The abstract paragraph should be indented \nicefrac{1}{2}~inch (3~picas) on both the left- and right-hand margins. Use 10~point type, with a vertical spacing (leading) of 11~points. The word \textbf{Abstract} must be centered, bold, and in point size 12. Two line spaces precede the abstract. The abstract must be limited to one paragraph.
\end{abstract}

\section{Method Details}
\label{sec:supp_method_details}

\subsection{Stage 1 Reward Design: Planning and Reflection Rewards}

During Stage 1, the agent acquires fundamental planning and diagnostic capabilities through supervised fine-tuning (SFT) and initial reinforcement learning. The reward signals during this phase are applied at a single-turn level.

\textbf{Planning Rewards ($R_{\mathrm{plan}}$)} comprise four verifiable metrics designed to evaluate the generated edit agenda:

\begin{equation}
    R_{\mathrm{plan}} = w_1 R_{\mathrm{JSON}} + w_2 R_{\mathrm{order}} + w_3 R_{\mathrm{type}} + w_4 R_{\mathrm{bbox}}
\end{equation}

where:
\begin{itemize}
    \item $R_{\mathrm{JSON}}$ verifies that the generated edit agenda is strictly parsable and accurately populates all requisite structural fields.
    
    \item $R_{\mathrm{order}}$ evaluates the logical sequencing of operations, heavily penalizing illogical temporal sequences 
    (e.g., executing an \op{add} operation before a prerequisite \op{remove} operation).
    
    \item $R_{\mathrm{type}}$ assesses the accuracy of the predicted operation types 
    (e.g., \op{move}, \op{add}, \op{remove}) against the ground truth.
    
    \item $R_{\mathrm{bbox}}$ measures the spatial alignment of the predicted source and target bounding boxes. 
    This is quantified using the Intersection over Union (IoU) between the predicted bounding box $B_p$ and the ground-truth box $B_{gt}$:
\end{itemize}

\begin{equation}
    \text{IoU}(B_p, B_{gt}) = \frac{\text{Area}(B_p \cap B_{gt})}{\text{Area}(B_p \cup B_{gt})}
\end{equation}

The coefficients $w_1$, $w_2$, $w_3$, and $w_4$ are hyperparameters that balance the contribution of each reward component. The default weights are set as $w_1 = 0.3$, $w_2 = 0.2$, $w_3 = 0.2$, and $w_4 = 0.3$.

\textbf{Reflection Rewards ($R_{\mathrm{reflect}}$)} evaluate the agent's diagnostic phase using three task-oriented signals:

\begin{equation}
R_{\mathrm{reflect}} =  w_5 R_{\mathrm{format}} + w_6 R_{\mathrm{status}} + w_7 R_{\mathrm{next}} 
\end{equation}

where:
\begin{itemize}
    \item $R_{\mathrm{format}}$ ensures that the structured diagnostic reflection strictly adheres to the required JSON schema.
    
    \item $R_{\mathrm{status}}$ evaluates the accuracy of the predicted intermediate state classification 
    (\op{success}, \op{continue}, or \op{rollback}).
    
    \item $R_{\mathrm{next}}$ assesses whether the agent correctly identifies the remaining task IDs when the current status is not \op{success}.
\end{itemize}

The coefficients $w_5$, $w_6$, and $w_7$ are hyperparameters that balance the contribution of each reward component. The default weights are set as $ w_5 = 0.4$, $w_6 = 0.3$, and $w_7 = 0.3$.

Additionally, we incorporate Chain-of-Thought (CoT) reasoning during the reflection phase to explicitly guide the agent through intermediate logical deliberation before outputting the final state diagnosis.

\subsection{Stage 3 Reward Design: Multi-turn Joint Reinforcement Training Rewards}

Stage 3 is the primary multi-turn training stage, where the agent interacts with the frozen mask-conditioned editor in a multi-turn loop. The rewards in this stage are dynamic and trajectory-based, focusing on the convergence of the entire editing session. In this multi-turn setting, the reward mechanism is bipartite, consisting of an initial planning reward and subsequent reflection rewards guided by a stronger VLM.

\textbf{Planning Reward.} During the initial turn of the editing trajectory, the agent generates the structured edit agenda. To maintain the foundational planning capabilities established earlier, the planning reward $R_{\mathrm{plan}}$ remains strictly consistent with the formulation in Stage 1. This ensures that the generated agenda maintains rigorous structural, sequential, and spatial alignment with the ground truth before execution begins.

\textbf{VLM-Guided Reflection Rewards.} For subsequent turns, we employ a stronger, more capable VLM as a Reward Model (RM) to evaluate the intermediate edited images. The RM independently assesses the current editing status after each editor execution. The reflection reward is then dynamically calculated based on the consistency between the agent's self-diagnosis and the RM's objective evaluation.

Specifically, the status classification reward $R_{\mathrm{status}}$ prioritizes "reliable convergence" over "greedy completion" through the following criteria:
\begin{itemize}
    \item \textbf{Success Reward (High Positive):} Awarded only when both the agent and the RM agree that all tasks are completed correctly.
    \item \textbf{Continue/Progress Reward (Cumulative Positive): }To encourage safe, incremental edits, the agent receives positive reinforcement for correctly identifying a \op{continue} state that aligns with the RM's assessment. This allows the system to build on partial successes rather than failing a complex request in one go.
    \item \textbf{Rollback Penalty (Heavy Negative):} If the agent triggers a \op{rollback}, it incurs a "rollback tax"—a penalty for inefficient execution. However, a "correct rollback" (where the agent identifies a fatal error corroborated by the RM and chooses to restart) is penalized far less severely than a "false success," thereby prioritizing product integrity.
    \item \textbf{Early Termination Penalty (Extreme Negative):} We strictly penalize the agent if it erroneously predicts \op{success} when the RM detects missed operations or visual corruption. This mechanism is crucial to prevent the model from exploiting a shortcut by indiscriminately predicting success to terminate the loop early.
\end{itemize}

Beyond these status-based rewards, the reflection phase strictly retains the format reward ($R_{\mathrm{format}}$) and the next-task identification reward ($R_{\mathrm{next}}$) introduced in Stage 1. This ensures that the structured diagnosis remains parsable and logically consistent for directing the subsequent turns.

\textbf{Trajectory-Level Advantage Allocation.} Unlike standard RL, the multi-turn reward is assigned to specific token spans across different turns. For a trajectory with $T$ rounds, the total reward is a combination of the initial planning quality (Turn 1) and the subsequent VLM-guided reflection accuracy (Turns 2 to $T$). We utilize a step-level advantage calculation to ensure that the reward from Round $t$ specifically reinforces the policy parameters responsible for that exact turn's output, preventing negative reflection rewards from degrading competent planning behaviors.

\section{Details of the Data Pipeline}

To construct a high-quality, fine-grained dataset for e-commerce image editing, our editable element mining pipeline leverages a suite of state-of-the-art vision and language models. This multi-stage perception and generation pipeline is designed to accurately decompose complex e-commerce posters into independent, structured semantic assets. The specific models and their corresponding roles are detailed below:

\textbf{Product and Graphic Object Localization: }We employ the YOLOv10~\cite{wang2024yolov10} object detector to identify and localize primary and salient visual elements within the source images. This model is responsible for generating precise bounding boxes for product instances, brand logos, decorative stickers, and other foreground graphic objects, serving as the foundational spatial prior for subsequent extraction.
    
\textbf{Layer Segmentation and Matting: }To ensure the extracted visual assets are of high fidelity and seamlessly reusable, we utilize a segmenter based on SAM2~\cite{ravi2024sam}. Conditioned on the bounding boxes provided by the localization stage, SAM2 performs pixel-level semantic segmentation and high-quality alpha matting. This process isolates the objects from complex backgrounds, yielding sharp, transparent masks that are crucial for downstream operations like \textit{move} and \textit{exchange}.

\textbf{Text and Layout Detection:} Recognizing the critical importance of text in e-commerce images, we rely on PP-OCRv5~\cite{cui2026pp} for comprehensive text analysis. This robust OCR model extracts both the textual content and the geometric layout of all text blocks. Capturing this text-specific metadata is essential for enabling localized, attribute-level modifications.

\textbf{Background Recovery: }Once the foreground elements (products, graphics, and text) are extracted, the remaining image contains missing regions. To create a clean, element-free background canvas, we deploy Flux2-klein-9B~\cite{blackforestlabs2025flux2}, a powerful diffusion-based inpainting model. This model intelligently fills the voids by referencing the surrounding context, resulting in a natural, semantically consistent background that serves as the foundation for generative recomposition.
    
\textbf{Structural Synthesis:} Finally, we utilize Qwen3.5~\cite{team2026qwen3}, a highly capable Multimodal Large Language Model, as the central integration engine. Acting as the cognitive orchestrator, Qwen3.5 synthesizes the parsed visual attributes, extracted textual content, and layout coordinates from the preceding modules. It translates these discrete, multi-modal components into a structured, machine-interpretable format (e.g., HTML). This hierarchical representation captures the underlying semantic logic and spatial constraints of the original poster, effectively forming the grounded "edit program" used to guide the training of both the VLM agent and the mask-conditioned editor.

\section{Implementation Details}

Our image editing backbone is initialized from Flux2-Klein-9B~\cite{blackforestlabs2025flux2}, while the VLM editing agent is based on Qwen3.5-9B~\cite{team2026qwen3}. The VLM handles instruction parsing, region-level planning, and visual reflection, whereas the editor receives the source image along with operation-aware masks and optional patch references. We optimize both models using AdamW. In Stage 1, the VLM undergoes supervised fine-tuning (SFT) with LoRA of rank 32 and a learning rate of $1\times10^{-4}$ for 20K steps with a batch size of 8. The Stage 1 GRPO training uses a group size of 8, learning rate $5\times10^{-6}$, batch size 16, and runs for 1,200 steps. The image editor is fine-tuned separately with LoRA of rank 128, learning rate $1\times10^{-4}$, for 25K steps. Stage 3 employs multi-turn GRPO with a sampling group of 8, LoRA rank 16, batch size 8, learning rate $1\times10^{-5}$, and 500 training steps. Qwen3-VL-32B~\ref{bai2025qwen3} serves as the reward model, and the maximum number of execution rounds during sampling is set to 4. Unless otherwise specified, all experiments are conducted on the same constructed training dataset and evaluated on the held-out \bench split.

\subsection{Style}

Papers to be submitted to NeurIPS 2026 must be prepared according to the
instructions presented here. Papers may only be up to {\bf nine} pages long, including figures. \textbf{Papers that exceed the page limit will not be reviewed (or in any other way considered) for presentation at the conference.}
Additional pages \emph{containing acknowledgments, references, checklist, and optional technical appendices} do not count as content pages.

The margins in 2026 are the same as those in previous years.

Authors are required to use the NeurIPS \LaTeX{} style files obtainable at the NeurIPS website as indicated below. Please make sure you use the current files and not previous versions. Tweaking the style files may be grounds for desk rejection.

\subsection{Retrieval of style files}

The style files for NeurIPS and other conference information are available on the website at
\begin{center}
  \url{https://neurips.cc}.
\end{center}

The only supported style file for NeurIPS 2026 is \verb+neurips_2026.sty+, rewritten for \LaTeXe{}. \textbf{Previous style files for \LaTeX{} 2.09,  Microsoft Word, and RTF are no longer supported.}

The \LaTeX{} style file contains three optional arguments: 
\begin{itemize}
\item \verb+final+, which creates a camera-ready copy, 
\item \verb+preprint+, which creates a preprint for submission to, e.g., arXiv, \item \verb+nonatbib+, which will not load the \verb+natbib+ package for you in case of package clash.
\end{itemize}

\paragraph{Preprint option}
If you wish to post a preprint of your work online, e.g., on arXiv, using the NeurIPS style, please use the \verb+preprint+ option. This will create a nonanonymized version of your work with the text ``Preprint. Work in progress.'' in the footer. This version may be distributed as you see fit, as long as you do not say which conference it was submitted to. Please \textbf{do not} use the \verb+final+ option, which should \textbf{only} be used for papers accepted to NeurIPS.

At submission time, please omit the \verb+final+ and \verb+preprint+ options. This will anonymize your submission and add line numbers to aid review. Please do \emph{not} refer to these line numbers in your paper as they will be removed during generation of camera-ready copies.

The file \verb+neurips_2026.tex+ may be used as a ``shell'' for writing your paper. All you have to do is replace the author, title, abstract, and text of the paper with your own.

The formatting instructions contained in these style files are summarized in Sections \ref{gen_inst}, \ref{headings}, and \ref{others} below.

\section{General formatting instructions}
\label{gen_inst}

The text must be confined within a rectangle 5.5~inches (33~picas) wide and
9~inches (54~picas) long. The left margin is 1.5~inch (9~picas).  Use 10~point
type with a vertical spacing (leading) of 11~points.  Times New Roman is the
preferred typeface throughout, and will be selected for you by default.
Paragraphs are separated by \nicefrac{1}{2}~line space (5.5 points), with no
indentation.

The paper title should be 17~point, initial caps/lower case, bold, centered
between two horizontal rules. The top rule should be 4~points thick and the
bottom rule should be 1~point thick. Allow \nicefrac{1}{4}~inch space above and
below the title to rules. All pages should start at 1~inch (6~picas) from the
top of the page.

For the final version, authors' names are set in boldface, and each name is
centered above the corresponding address. The lead author's name is to be listed
first (left-most), and the co-authors' names (if different address) are set to
follow. If there is only one co-author, list both author and co-author side by
side.

Please pay special attention to the instructions in Section \ref{others}
regarding figures, tables, acknowledgments, and references.

\section{Headings: first level}
\label{headings}

All headings should be lower case (except for first word and proper nouns),
flush left, and bold.

First-level headings should be in 12-point type.

\subsection{Headings: second level}

Second-level headings should be in 10-point type.

\subsubsection{Headings: third level}

Third-level headings should be in 10-point type.

\paragraph{Paragraphs}

There is also a \verb+\paragraph+ command available, which sets the heading in
bold, flush left, and inline with the text, with the heading followed by 1\,em
of space.

\section{Citations, figures, tables, references}
\label{others}

These instructions apply to everyone.

\subsection{Citations within the text}

The \verb+natbib+ package will be loaded for you by default.  Citations may be
author/year or numeric, as long as you maintain internal consistency.  As to the
format of the references themselves, any style is acceptable as long as it is
used consistently.

The documentation for \verb+natbib+ may be found at
\begin{center}
  \url{http://mirrors.ctan.org/macros/latex/contrib/natbib/natnotes.pdf}
\end{center}
Of note is the command \verb+\citet+, which produces citations appropriate for
use in inline text.  For example,
\begin{verbatim}
   \citet{hasselmo} investigated\dots
\end{verbatim}
produces
\begin{quote}
  Hasselmo, et al.\ (1995) investigated\dots
\end{quote}

If you wish to load the \verb+natbib+ package with options, you may add the
following before loading the \verb+neurips_2026+ package:
\begin{verbatim}
   \PassOptionsToPackage{options}{natbib}
\end{verbatim}

If \verb+natbib+ clashes with another package you load, you can add the optional
argument \verb+nonatbib+ when loading the style file:
\begin{verbatim}
   \usepackage[nonatbib]{neurips_2026}
\end{verbatim}

As submission is double blind, refer to your own published work in the third
person. That is, use ``In the previous work of Jones et al.\ [4],'' not ``In our
previous work [4].'' If you cite your other papers that are not widely available
(e.g., a journal paper under review), use anonymous author names in the
citation, e.g., an author of the form ``A.\ Anonymous'' and include a copy of the anonymized paper in the supplementary material.

\subsection{Footnotes}

Footnotes should be used sparingly.  If you do require a footnote, indicate
footnotes with a number\footnote{Sample of the first footnote.} in the
text. Place the footnotes at the bottom of the page on which they appear.
Precede the footnote with a horizontal rule of 2~inches (12~picas).

Note that footnotes are properly typeset \emph{after} punctuation
marks.\footnote{As in this example.}

\subsection{Figures}

\begin{figure}
  \centering
  \fbox{\rule[-.5cm]{0cm}{4cm} \rule[-.5cm]{4cm}{0cm}}
  \caption{Sample figure caption. Explain what the figure shows and add a key take-away message to the caption.}
\end{figure}

All artwork must be neat, clean, and legible. Lines should be dark enough for
 reproduction purposes. The figure number and caption always appear after the
figure. Place one line space before the figure caption and one line space after
the figure. The figure caption should be lower case (except for the first word and proper nouns); figures are numbered consecutively.

You may use color figures.  However, it is best for the figure captions and the
paper body to be legible if the paper is printed in either black/white or in
color.

\subsection{Tables}

All tables must be centered, neat, clean, and legible.  The table number and
title always appear before the table.  See Table~\ref{sample-table}.

Place one line space before the table title, one line space after the
table title, and one line space after the table. The table title must
be lower case (except for the first word and proper nouns); tables are
numbered consecutively.

Note that publication-quality tables \emph{do not contain vertical rules}. We
strongly suggest the use of the \verb+booktabs+ package, which allows for
typesetting high-quality, professional tables:
\begin{center}
  \url{https://www.ctan.org/pkg/booktabs}
\end{center}
This package was used to typeset Table~\ref{sample-table}.

\begin{table}
  \caption{Sample table caption. Explain what the table shows and add a key take-away message to the caption.}
  \label{sample-table}
  \centering
  \begin{tabular}{lll}
    \toprule
    \multicolumn{2}{c}{Part}                   \\
    \cmidrule(r){1-2}
    Name     & Description     & Size ($\mu$m) \\
    \midrule
    Dendrite & Input terminal  & $\approx$100     \\
    Axon     & Output terminal & $\approx$10      \\
    Soma     & Cell body       & up to $10^6$  \\
    \bottomrule
  \end{tabular}
\end{table}

\subsection{Math}
Note that display math in bare TeX commands will not create correct line numbers for submission. Please use LaTeX (or AMSTeX) commands for unnumbered display math. (You really shouldn't be using \$\$ anyway; see \url{https://tex.stackexchange.com/questions/503/why-is-preferable-to} and \url{https://tex.stackexchange.com/questions/40492/what-are-the-differences-between-align-equation-and-displaymath} for more information.)

\subsection{Final instructions}

Do not change any aspects of the formatting parameters in the style files.  In
particular, do not modify the width or length of the rectangle the text should
fit into, and do not change font sizes. Please note that pages should be
numbered.

\section{Preparing PDF files}

Please prepare submission files with paper size ``US Letter,'' and not, for
example, ``A4.''

Fonts were the main cause of problems in the past years. Your PDF file must only
contain Type 1 or Embedded TrueType fonts. Here are a few instructions to
achieve this.

\begin{itemize}

\item You should directly generate PDF files using \verb+pdflatex+.

\item You can check which fonts a PDF files uses.  In Acrobat Reader, select the
  menu Files$>$Document Properties$>$Fonts and select Show All Fonts. You can
  also use the program \verb+pdffonts+ which comes with \verb+xpdf+ and is
  available out-of-the-box on most Linux machines.

\item \verb+xfig+ ``patterned'' shapes are implemented with bitmap fonts.  Use
  "solid" shapes instead.

\item The \verb+\bbold+ package almost always uses bitmap fonts.  You should use
  the equivalent AMS Fonts:
\begin{verbatim}
   \usepackage{amsfonts}
\end{verbatim}
followed by, e.g., \verb+\mathbb{R}+, \verb+\mathbb{N}+, or \verb+\mathbb{C}+
for $\mathbb{R}$, $\mathbb{N}$ or $\mathbb{C}$.  You can also use the following
workaround for reals, natural and complex:
\begin{verbatim}
   \newcommand{\RR}{I\!\!R} %real numbers
   \newcommand{\Nat}{I\!\!N} %natural numbers
   \newcommand{\CC}{I\!\!\!\!C} %complex numbers
\end{verbatim}
Note that \verb+amsfonts+ is automatically loaded by the \verb+amssymb+ package.

\end{itemize}

If your file contains type 3 fonts or non embedded TrueType fonts, we will ask
you to fix it.

\subsection{Margins in \LaTeX{}}

Most of the margin problems come from figures positioned by hand using
\verb+\special+ or other commands. We suggest using the command
\verb+\includegraphics+ from the \verb+graphicx+ package. Always specify the
figure width as a multiple of the line width as in the example below:
\begin{verbatim}
   \usepackage[pdftex]{graphicx} ...
   \includegraphics[width=0.8\linewidth]{myfile.pdf}
\end{verbatim}
See Section 4.4 in the graphics bundle documentation
(\url{http://mirrors.ctan.org/macros/latex/required/graphics/grfguide.pdf})

A number of width problems arise when \LaTeX{} cannot properly hyphenate a
line. Please give LaTeX hyphenation hints using the \verb+\-+ command when
necessary.

\begin{ack}
Use unnumbered first level headings for the acknowledgments. All acknowledgments
go at the end of the paper before the list of references. Moreover, you are required to declare
funding (financial activities supporting the submitted work) and competing interests (related financial activities outside the submitted work).
More information about this disclosure can be found at: \url{https://neurips.cc/Conferences/2026/PaperInformation/FundingDisclosure}.

Do {\bf not} include this section in the anonymized submission, only in the final paper. You can use the \texttt{ack} environment provided in the style file to automatically hide this section in the anonymized submission.
\end{ack}

\section*{References}

References follow the acknowledgments in the camera-ready paper. Use unnumbered first-level heading for
the references. Any choice of citation style is acceptable as long as you are
consistent. It is permissible to reduce the font size to \verb+small+ (9 point)
when listing the references.
Note that the Reference section does not count towards the page limit.
\medskip

{
\small

[1] Alexander, J.A.\ \& Mozer, M.C.\ (1995) Template-based algorithms for
connectionist rule extraction. In G.\ Tesauro, D.S.\ Touretzky and T.K.\ Leen
(eds.), {\it Advances in Neural Information Processing Systems 7},
pp.\ 609--616. Cambridge, MA: MIT Press.

[2] Bower, J.M.\ \& Beeman, D.\ (1995) {\it The Book of GENESIS: Exploring
  Realistic Neural Models with the GEneral NEural SImulation System.}  New York:
TELOS/Springer--Verlag.

[3] Hasselmo, M.E., Schnell, E.\ \& Barkai, E.\ (1995) Dynamics of learning and
recall at excitatory recurrent synapses and cholinergic modulation in rat
hippocampal region CA3. {\it Journal of Neuroscience} {\bf 15}(7):5249-5262.
}


\appendix

\section{Technical appendices and supplementary material}
Technical appendices with additional results, figures, graphs, and proofs may be submitted with the paper submission before the full submission deadline (see above). You can upload a ZIP file for videos or code, but do not upload a separate PDF file for the appendix. There is no page limit for the technical appendices. 

Note: Think of the appendix as ``optional reading'' for reviewers. The paper must be able to stand alone without the appendix; for example, adding critical experiments that support the main claims to an appendix is inappropriate. 


\newpage
\section*{NeurIPS Paper Checklist}

\begin{enumerate}

\item {\bf Claims}
    \item[] Question: Do the main claims made in the abstract and introduction accurately reflect the paper's contributions and scope?
    \item[] Answer: \answerYes{} 
    \item[] Justification: The abstract and introduction outline our theoretical and technical contributions, the proposed methodology, and a comprehensive experimental evaluation.

\item {\bf Limitations}
    \item[] Question: Does the paper discuss the limitations of the work performed by the authors?
    \item[] Answer: \answerYes{} 
    \item[] Justification: At the end of the appendix, a limitation discussion is provided. 

\item {\bf Theory assumptions and proofs}
    \item[] Question: For each theoretical result, does the paper provide the full set of assumptions and a complete (and correct) proof?
    \item[] Answer: \answerYes{} 
    \item[] Justification: A complete proof is given in the method section.

    \item {\bf Experimental result reproducibility}
    \item[] Question: Does the paper fully disclose all the information needed to reproduce the main experimental results of the paper to the extent that it affects the main claims and/or conclusions of the paper (regardless of whether the code and data are provided or not)?
    \item[] Answer: \answerYes{} 
    \item[] Justification: Yes, the baseline model, technical details, hyperparameters, and configuration are detailed in the submission for reproducibility.

\item {\bf Open access to data and code}
    \item[] Question: Does the paper provide open access to the data and code, with sufficient instructions to faithfully reproduce the main experimental results, as described in supplemental material?
    \item[] Answer: \answerYes{} 
    \item[] Justification: The datasets this paper uses are publicly available, and the source code is promised to be public once published.

\item {\bf Experimental setting/details}
    \item[] Question: Does the paper specify all the training and test details (e.g., data splits, hyperparameters, how they were chosen, type of optimizer) necessary to understand the results?
    \item[] Answer: \answerYes{} 
    \item[] Justification: The implementation details are given at the beginning of the experimental section and the beginning of the appendix.

\item {\bf Experiment statistical significance}
    \item[] Question: Does the paper report error bars suitably and correctly defined or other appropriate information about the statistical significance of the experiments?
    \item[] Answer: \answerNo{} 
    \item[] Justification: Following prior works in this field, the evaluation protocols on the corresponding datasets do NOT require a report of the error bar.

\item {\bf Experiments compute resources}
    \item[] Question: For each experiment, does the paper provide sufficient information on the computer resources (type of compute workers, memory, time of execution) needed to reproduce the experiments?
    \item[] Answer: \answerYes{} 
    \item[] Justification: The implementation details are given at the beginning of the experimental section and the beginning of the appendix.
    
\item {\bf Code of ethics}
    \item[] Question: Does the research conducted in the paper conform, in every respect, with the NeurIPS Code of Ethics \url{https://neurips.cc/public/EthicsGuidelines}?
    \item[] Answer: \answerYes{} 
    \item[] Justification: This paper focuses on a fundamental task of machine learning and conducts experiments on publicly available datasets.

\item {\bf Broader impacts}
    \item[] Question: Does the paper discuss both potential positive societal impacts and negative societal impacts of the work performed?
    \item[] Answer: \answerYes{} 
    \item[] Justification: The societal impact of this work has been discussed at the end of the conclusion section. We do not envision negative societal impact could be brought by this work.
    
\item {\bf Safeguards}
    \item[] Question: Does the paper describe safeguards that have been put in place for responsible release of data or models that have a high risk for misuse (e.g., pre-trained language models, image generators, or scraped datasets)?
    \item[] Answer: \answerNA{} 
    \item[] Justification: We do not envision such risks.

\item {\bf Licenses for existing assets}
    \item[] Question: Are the creators or original owners of assets (e.g., code, data, models), used in the paper, properly credited and are the license and terms of use explicitly mentioned and properly respected?
    \item[] Answer: \answerYes{} 
    \item[] Justification: All the assets have been properly cited, with a license to use for academia and no commercial purpose.

\item {\bf New assets}
    \item[] Question: Are new assets introduced in the paper well documented and is the documentation provided alongside the assets?
    \item[] Answer: \answerYes{} 
    \item[] Justification: We will release new assets.

\item {\bf Crowdsourcing and research with human subjects}
    \item[] Question: For crowdsourcing experiments and research with human subjects, does the paper include the full text of instructions given to participants and screenshots, if applicable, as well as details about compensation (if any)? 
    \item[] Answer: \answerNA{} 
    \item[] Justification: This paper does not involve crowdsourcing nor research with human subjects.

\item {\bf Institutional review board (IRB) approvals or equivalent for research with human subjects}
    \item[] Question: Does the paper describe potential risks incurred by study participants, whether such risks were disclosed to the subjects, and whether Institutional Review Board (IRB) approvals (or an equivalent approval/review based on the requirements of your country or institution) were obtained?
    \item[] Answer: \answerNA{} 
    \item[] Justification: This paper does not involve crowdsourcing nor research with human subjects

\item {\bf Declaration of LLM usage}
    \item[] Question: Does the paper describe the usage of LLMs if it is an important, original, or non-standard component of the core methods in this research? Note that if the LLM is used only for writing, editing, or formatting purposes and does \emph{not} impact the core methodology, scientific rigor, or originality of the research, declaration is not required.
    \item[] Answer: \answerYes{} 
    \item[] Justification: We leverage Large Language Models (LLMs) for both data curation and evaluation.

\end{enumerate}